\documentclass[letterpaper,journal,twoside]{IEEEtran}
\IEEEoverridecommandlockouts                             
\usepackage{graphicx}
\usepackage{xurl}    
\usepackage{amsmath,amssymb}
\usepackage{placeins}
\usepackage{dblfloatfix}
\usepackage{multirow}
\usepackage{xcolor}
\usepackage{colortbl}
\usepackage{pifont} 
\usepackage{booktabs}
\definecolor{avggray}{gray}{0.92}
\usepackage{subfig}
\usepackage{cite}

\usepackage[hidelinks]{hyperref}

\title{\LARGE \bf
Zero-Shot Cross-City Generalization in End-to-End Autonomous Driving: Self-Supervised versus Supervised Representations}

\author{Fatemeh Naeinian, Ali Hamza, Haoran Zhu, and Anna Choromanska
\thanks{Department of Electrical and Computer Engineering, NYU Tandon School of Engineering, Brooklyn, NY, USA.}%
\thanks{Email: \{fn2174, ah7072, hz1922, ac5455\}@nyu.edu}%
}

\begin{document}

\maketitle
\thispagestyle{empty}
\pagestyle{empty}

\begin{abstract}
End-to-end autonomous driving models are typically trained on multi-city datasets using supervised ImageNet-pretrained backbones, yet their ability to generalize to unseen cities remains largely unexamined. When training and evaluation data are geographically mixed, models may implicitly rely on city-specific cues, masking failure modes that would occur under real-world domain shifts when generalizing to new locations. In this work, we formulate zero-shot cross-city transfer as a controlled representation-level stress test for end-to-end autonomous driving and ask how visual pretraining affects transfer behavior under geographic domain shift. We conduct a comprehensive study by integrating self-supervised backbones I-JEPA, DINOv2, and MAE into planning frameworks. We evaluate performance under strict geographic splits on nuScenes in the open-loop setting and on NAVSIM using a pseudo-simulation evaluation protocol. Our experiments reveal a substantial generalization gap when transferring models across cities with different road topologies, traffic conventions, and visual environments. In open-loop evaluation, a supervised backbone exhibits severe degradation when transferring between cities, yet a domain-specific self-supervised approach can substantially reduce both displacement and collision degradation. Also, in pseudo-simulation evaluation, self-supervised pretraining improves average out-of-distribution EPDMS in several single-city training settings. Our results provide empirical evidence that representation learning influences the robustness of cross-city planning and motivate zero-shot geographic transfer as an important stress test for evaluating end-to-end autonomous driving systems.
\end{abstract}

\begin{IEEEkeywords}
Autonomous driving, computer vision,
self-supervised learning, geographic generalization.
\end{IEEEkeywords}
\section{\textbf{Introduction}}

Autonomous driving systems must operate reliably across cities with different road structures, traffic conventions, and environmental characteristics. However, many end-to-end autonomous driving models are trained and evaluated on geographically mixed datasets, where training and validation samples have very similar urban distributions. Under such settings, performance metrics primarily reflect interpolation within a known distribution rather than true generalization. In real-world deployment, a vehicle trained in one city must operate safely in another without access to city-specific retraining. Collecting and annotating data for every possible city or deployment location is economically and operationally infeasible, making zero-shot cross-city generalization an important requirement for scalable autonomy.

Zero-shot cross-city transfer presents a more realistic and challenging evaluation scenario. In this setting, a model is trained exclusively on data from one city and evaluated on a geographically distinct city. Differences in road topology, driving orientation (e.g., right-hand versus left-hand traffic), traffic density, infrastructure design, and visual statistics introduce structural domain shifts that challenge learned representations. As our experiments show, performance can degrade drastically under such conditions; most prior work reports aggregate results on mixed splits without isolating the effect of geographic domain shift. As a result, the robustness of end-to-end planners under explicit city-level shift remains insufficiently understood.

End-to-end autonomous driving systems \cite{chen2024end,hu2023planningorientedautonomousdriving} are particularly sensitive to this issue. Unlike modular pipelines, which rely on intermediate detection or mapping outputs from separate networks within each module, end-to-end systems rely on a shared backbone that serves as a feature extractor, encoding raw sensory inputs, such as multi-view images (optionally LiDAR and radar), into a latent space. A planning head is then attached to map this latent representation directly to future trajectories. These systems may optionally include auxiliary perception tasks to inject stronger inductive bias; nevertheless, they still use a shared backbone as a general feature extractor. Therefore, the robustness of the backbone's learned representations is crucial. If the latent features encode city-specific biases, the resulting planner may fail substantially when deployed in unseen environments.

This motivates the central question of this work:

\begin{quote}
\textit{\textbf{How does representation pretraining affect zero-shot cross-city generalization in end-to-end autonomous driving?}}
\end{quote}

Rather than treating cross-city evaluation as another benchmark split, we use it as a
controlled representation-level stress test. In a standard mixed-city evaluation, training and
validation data may share similar geographic cues, allowing models to achieve strong
aggregate performance while relying on city-specific regularities. By separating training
and evaluation cities, we test whether the learned visual representation supports
transferable driving behavior or instead encodes brittle geographic biases. This framing
allows us to study not only whether a backbone improves average accuracy, but also how it
changes the structure of failure under transfer, including the magnitude and directionality
of degradation across cities.

To instantiate this stress test, we use geographically separated splits of
nuScenes~\cite{caesar2020nuscenesmultimodaldatasetautonomous} and
NAVSIM~\cite{dauner2024navsimdatadrivennonreactiveautonomous}. We train end-to-end
autonomous driving models exclusively on one city and evaluate them zero-shot on unseen cities,
without adaptation. Within this protocol, we compare
supervised ImageNet-pretrained backbones, generic self-supervised backbones, and
domain-specific self-supervised backbones pretrained on driving data. Our goal is not only
to improve benchmark scores, but to analyze how representation pretraining affects
robustness under geographic domain shift.

Representation learning has recently become a dominant paradigm in computer vision. Large-scale supervised pretraining (e.g., on ImageNet~\cite{russakovsky2015imagenetlargescalevisual}) and self-supervised learning (SSL) methods such as I-JEPA~\cite{assran2023selfsupervisedlearningimagesjointembedding}, DINOv2~\cite{oquab2024dinov2learningrobustvisual}, and MAE~\cite{he2021maskedautoencodersscalablevision} have demonstrated strong transferability across downstream tasks. Recent work has also shown that such representations can achieve high performance in end-to-end autonomous driving under standard mixed-city training settings~\cite{wang2026drivejepavideojepameets}.
However, their impact on structural domain shift in trajectory planning remains unclear. In particular, it is unknown whether stronger visual representations merely improve
in-distribution accuracy or whether they also change how planners fail under cross-city
transfer.

The contributions of this work are threefold:
(i) We formulate zero-shot cross-city transfer as a controlled representation-level stress
test for end-to-end autonomous driving, isolating geographic generalization from standard
mixed-city interpolation;
(ii) We reveal that mixed-city evaluation can mask severe directional failures, where a model trained in one city may fail substantially more when transferred to another city than
in the reverse direction; and
(iii) We show that visual representation pretraining affects not only average
planning accuracy, but also the structure of transfer behavior under domain
shift, with domain-specific self-supervised representations reducing transfer
degradation in selected settings.

\begin{figure*}[t]
    \centering
    \includegraphics[
        pagebox=cropbox,
        width=\textwidth,
        height=0.55\textheight,
        keepaspectratio
    ]{fig/framework4.pdf}
    \caption{
    Overview of the evaluation framework.
    Left: region-specific self-supervised pretraining of I-JEPA, DINOv2, and
    MAE on balanced NAVSIM NavTrain imagery.
    Center: integration of the pretrained visual backbones into LAW and
    TransFuser.
    Right: zero-shot geographic evaluation on unseen cities or regions without
    target-domain adaptation.
    }
    \label{fig:framework}
\end{figure*}

\section{\textbf{RELATED WORK}}
\subsection{\textbf{End-to-End Autonomous Driving and World Models}}

Early end-to-end autonomous driving systems~\cite{pomerleau1988alvinn, bojarski2016endendlearningselfdriving} directly map sensor inputs to control commands without explicit modular decomposition. Imitation learning has since
become a widely studied paradigm for end-to-end autonomous
driving~\cite{9700770}, with subsequent work extending these systems to
multimodal sensor inputs~\cite{Multimodal}. Data-driven simulation has also been used to train robust perception-to-control policies that transfer to unseen real roads~\cite{8957584}. Modern approaches~\cite{chen2020end} instead aim to jointly model perception, prediction, and planning within unified architectures that output future ego trajectories or control signals. Representative examples include VAD~\cite{jiang2023vad} and UniAD~\cite{hu2023planningorientedautonomousdriving}, which integrate multi-task reasoning into structured planning pipelines. More recent planning-centric systems, including LAW~\cite{li2024enhancing}, SSR~\cite{li2025navigationguidedsparsescenerepresentation}, WoTE~\cite{li2025endtoenddrivingonlinetrajectory}, World4Drive~\cite{zheng2025world4driveendtoendautonomousdriving}, TransFuser~\cite{chitta2022TransFuserimitationtransformerbasedsensor}, Hydra-MDP~\cite{li2024hydramdpendtoendmultimodalplanning}, SeerDrive~\cite{zhang2025futureawareendtoenddrivingbidirectional}, SparseDrive~\cite{sun2024sparsedriveendtoendautonomousdriving}, ComDrive~\cite{wang2025comdrivecomfortorientedendtoendautonomous}, iPad~\cite{guo2025ipaditerativeproposalcentricendtoend}, and METDrive~\cite{guo2025metdrive}, report strong in-domain performance by incorporating multimodal fusion, structured scene representation, trajectory-centric supervision, and joint future generation and perception. Related work also improves E2E driving through BEV-based interpretable imitation learning~\cite{11373828} and learned continuous planning costs for safer, more road-compliant behavior~\cite{11393633}. Among these approaches, LAW explicitly models action-conditioned future latent dynamics, encouraging temporally consistent world representations, making it particularly suitable for studying how visual representation quality influences downstream planning behavior.

Large-scale datasets, such as nuScenes, enable open-loop evaluation, while other benchmarks, such as NAVSIM, support pseudo-simulation assessment. Together, these benchmarks enable us to analyze the interactions among representation quality, learned planning dynamics, and zero-shot cross-city generalization. Recent work shows that pseudo-simulation better captures error recovery and correlates more strongly with closed-loop driving behavior than standard open-loop metrics~\cite{cao2025pseudosimulationautonomousdriving},
and pseudo-simulation-based reinforcement learning has been proposed to address the misalignment between imitation-learning objectives and closed-loop driving requirements~\cite{perlad}. Similarly, nuScenes open-loop evaluation can be misleading because models may rely heavily on ego-status shortcuts rather than perception; ego-status-based models can perform competitively under standard L2 and collision metrics while hiding perception failures~\cite{Li_2024_CVPR}. Therefore, we report nuScenes open-loop L2 and collision rate together with NAVSIM pseudo-simulation-based EPDMS for a more complete evaluation.

\subsection{\textbf{Supervised Visual Backbones}}
Visual backbones are central to end-to-end autonomous driving systems, as they determine the quality of features used for prediction and planning. Early approaches relied on supervised pretraining on large-scale datasets such as ImageNet~\cite{russakovsky2015imagenetlargescalevisual}, commonly using convolutional architectures like ResNet~\cite{he2015deepresiduallearningimage} for their strong spatial inductive biases and stable optimization. More recently, transformer-based encoders such as Vision Transformers~\cite{dosovitskiy2021imageworth16x16words} and Swin Transformer~\cite{liu2021swintransformerhierarchicalvision} have gained popularity. Finally, BEV-centric models like BEVFormer~\cite{li2022bevformerlearningbirdseyeviewrepresentation} explicitly model spatial structure to better support downstream planning.

\subsection{\textbf{Self-Supervised Visual Backbones}}
Self-supervised learning (SSL) has emerged as a key paradigm for visual representation learning, enabling large-scale pretraining without manual annotations. SSL foundation models such as DINOv2~\cite{oquab2024dinov2learningrobustvisual} and MAE~\cite{he2021maskedautoencodersscalablevision} demonstrate that robust visual representations can be effectively transferred across downstream tasks. Recently, a new SSL paradigm based on joint-embedding predictive architectures, such as I-JEPA~\cite{assran2023selfsupervisedlearningimagesjointembedding}, V-JEPA~\cite{bardes2024revisiting}, and V-JEPA~2~\cite{assran2025vjepa2}, has emerged. JEPA enables learning compact, semantically meaningful features by predicting masked or future latent representations rather than reconstructing pixels, thereby enabling efficient video pretraining at scale. AD-L-JEPA~\cite{zhu2025selfsupervisedrepresentationlearningjoint} is the first JEPA-based method for the autonomous driving domain using LiDAR data. Subsequently, Drive-JEPA~\cite{wang2026drivejepavideojepameets} shows that a V-JEPA-pretrained backbone can significantly improve end-to-end planning performance. However, generalizing different SSL-pretrained backbones across multi-city autonomous driving settings remains unexplored.

\subsection{\textbf{Domain Generalization in Autonomous Driving}}

Domain shift remains a major challenge for autonomous driving systems deployed in unseen environments. Early generalization studies relied on simulation platforms such as CARLA~\cite{dosovitskiy2017carlaopenurbandriving}, but synthetic settings fail to capture real-world urban variability. Generalization has also been
studied in trajectory prediction; MetaTraj~\cite{MetaTraj} uses meta-learning
to improve transfer across scenes and object types. In real-world planning, methods like RoCA~\cite{yasarla2025rocarobustcrossdomainendtoend} improve cross-domain robustness through probabilistic modeling, though strong gains often require target-domain adaptation. Other approaches leverage LLM-based evaluation~\cite{dong2024generalizingendtoendautonomousdriving} or multi-modal foundation models~\cite{wang2023driveanywheregeneralizableendtoend} to enhance reasoning under distribution shifts. Recent work also shows that in-distribution leaderboard rankings may not reflect cross-dataset robustness~\cite{Yao2024ImprovingOG}, and that representation structure plays a critical role in maintaining performance under domain shift~\cite{singh2025mattersrepresentationalignmentglobal}.

\section{\textbf{METHODS}}

Our objective is to isolate the effect of self-supervised visual representation pretraining on zero-shot cross-city generalization of trajectory planning under controlled conditions. To this end, we design a three-stage experimental framework: (1) backbone pretraining using domain-specific self-supervised learning; (2) integration of the pretrained backbone into the end-to-end autonomous driving architectures; and (3) evaluation under strict geographic zero-shot splits in both open-loop and pseudo-simulation settings.  An overview of the studied evaluation framework is illustrated in Fig.~\ref{fig:framework}. Within each architecture, the planning module is held constant, ensuring that observed differences arise from representation initialization rather than downstream design choices.

\subsection{\textbf{End-to-End Planning Architectures}}
We evaluate cross-city generalization in both open-loop (nuScenes) and pseudo-simulation (NAVSIM) settings, where open-loop evaluation
compares predicted trajectories against logged ground truth, and NAVSIM evaluates
predicted trajectories using simulation-based driving metrics. For open-loop evaluation, we adopt LAW~\cite{li2024enhancing}, replacing its Swin-Transformer-Tiny backbone with a pretrained ViT. For pseudo-simulation evaluation, given the substantially larger dataset and computational cost, we use TransFuser~\cite{chitta2022TransFuserimitationtransformerbasedsensor}, replacing its ResNet34 image encoder with a pretrained ViT while keeping the LiDAR encoder unchanged. We also report results on Latent TransFuser~\cite{chitta2022TransFuserimitationtransformerbasedsensor}, an image-only variant that replaces the LiDAR branch with a fixed positional encoding.

\subsection{\textbf{Visual Backbone Variants}}

We evaluate three SSL pretrained visual backbones: I-JEPA, DINOv2, and MAE. Our goal is to compare generic large-scale pretrained representations against domain-specific self-supervised pretraining.

\textbf{Public Pretrained Backbones.}
As a reference baseline, we use publicly available pretrained Vision Transformers: I-JEPA ViT-H/14, DINOv2 ViT-g/14, and MAE ViT-H/14, integrated into LAW and TransFuser / Latent TransFuser. Because these models differ in capacity (e.g., ViT-H/14 vs.\ ViT-g/14), we treat them as system-level reference backbones rather than controlled comparisons of SSL objectives.

\textbf{Domain-Specific Self-Supervised Pretraining.}
To ensure a fair comparison across self-supervised learning objectives, we perform domain-specific pretraining on balanced NAVSIM NavTrain driving sequences using the same ViT-S/14 backbone configuration for all three methods (I-JEPA, DINOv2, and MAE). By keeping the architecture and capacity fixed, we control these factors when comparing the pretraining objectives. Each domain contributes $16,303$ three-camera frames, corresponding to $48,909$ images. Separate checkpoints are trained for Boston, Las Vegas, Pittsburgh, Singapore, the combined four-region setting, and Boston+Singapore. The same NAVSIM-pretrained checkpoints are used in LAW experiments.

\subsection{\textbf{Zero-Shot Cross-City Evaluation Protocol}}

We adopt strict geographic splits in which models are trained exclusively
on data from a single city and evaluated on unseen cities without any
fine-tuning or adaptation on the target domain.

\paragraph{\textbf{nuScenes}}
For open-loop evaluation, we use the nuScenes dataset, which contains real-world driving data collected in Boston and Singapore. We treat these cities as distinct geographic domains, with Boston representing a right-hand driving environment and Singapore representing a left-hand driving environment. We evaluate bidirectional transfer, Boston$\rightarrow$Singapore and Singapore$\rightarrow$Boston, and report the corresponding in-domain baselines. Boston contains 390 training and 77 validation scenes; Singapore contains 310 training and 73 validation scenes. The three Singapore geographic subregions are grouped into a single city-level domain.

\paragraph{\textbf{NAVSIM}}
For pseudo-simulation evaluation, we use the NAVSIM dataset and benchmark, which provide driving logs and pseudo-simulation-based evaluation across Boston, Pittsburgh, Las Vegas (right-hand driving), and Singapore (left-hand driving). We train on one city at a time using the corresponding NavTrain split and evaluate on the remaining cities using NavTest. NavTrain contains 818 logs in Las Vegas, 255 in Boston, 140 in Pittsburgh, and 97 in Singapore; NavTest contains 50, 62, 22, and 13 logs,
respectively. To remove training set size as a confounding factor, we construct
equal-sized regional training subsets.

This protocol isolates geographic domain shifts in traffic conventions, road topology, infrastructure, and visual appearance, allowing us to quantify the generalization gap between in-domain and cross-city evaluation. For single-region experiments, the SSL checkpoint is pretrained only on
imagery from the corresponding source region; no target-region imagery is
used during SSL pretraining. Joint four-region experiments use the
four-region checkpoint, while joint Boston+Singapore experiments use the
Boston+Singapore checkpoint.

\paragraph{\textbf{Data Balancing Protocol}}
We balance data for downstream planning and SSL pretraining independently because they
operate on windowed planning samples and synchronized camera frames, respectively.
For downstream training, we sample 10{,}821 examples per region, matching the
smallest NavTrain set (Pittsburgh). For SSL pretraining, we sample 16{,}303
eligible frames per region, matching Singapore; with three synchronized camera
views treated as independent images, this yields 48{,}909 images per region.
Table~\ref{tab:navtrain-scene-balance} summarizes the resulting balanced sets.

\paragraph{\textbf{Test-Time Horizontal-Flip Diagnostic}}
We additionally evaluate a diagnostic in which the camera images
are horizontally flipped at test time before inference. The model predicts
a trajectory from the flipped observations, and the predicted trajectory
is then horizontally flipped back to the original coordinate frame before
evaluation. Model parameters and all other aspects of the evaluation
protocol remain unchanged. This diagnostic is reported separately and is
not part of the standard zero-shot protocol.

\begin{table}[t]
    \centering
    \caption{
    NAVSIM NavTrain data volume and balanced subsets. Downstream training uses
    10{,}821 windowed samples per region. SSL pretraining uses 16{,}303
    eligible three-camera frames per region, corresponding to 48{,}909
    individual images. 
    }
    \label{tab:navtrain-scene-balance}
    \footnotesize
    \setlength{\tabcolsep}{2.5pt}
    \renewcommand{\arraystretch}{1.2}
    \begin{tabular}{lcccc}
    \toprule
    \textbf{Region}
    & \shortstack{\textbf{Available}\\\textbf{planning samples}}
    & \shortstack{\textbf{Planning}\\\textbf{used}}
    & \shortstack{\textbf{Eligible}\\\textbf{SSL frames}}
    & \shortstack{\textbf{SSL frames}\\\textbf{used}} \\
    \midrule
    Las Vegas   & 67{,}683 & 10{,}821 & 99{,}690 & 16{,}303 \\
    Boston      & 12{,}954 & 10{,}821 & 19{,}851 & 16{,}303 \\
    Pittsburgh  & 10{,}821 & 10{,}821 & 16{,}651 & 16{,}303 \\
    Singapore   & 11{,}830 & 10{,}821 & 16{,}303 & 16{,}303 \\
    \bottomrule
    \end{tabular}
\end{table}

\section{EXPERIMENTS}
\subsection{\textbf{Benchmarks and Metrics}}

\paragraph{\textbf{nuScenes}}
We evaluate open-loop end-to-end planning performance on the nuScenes dataset. This setting naturally enables the study of cross-city generalization under domain shift. Performance is measured using L2 displacement error (in meters) and collision rate (in \%) for the predicted ego-vehicle trajectory over a 3-second horizon, with waypoints sampled at 2~Hz.

\paragraph{\textbf{NAVSIM v2}}
To evaluate geographic transfer under pseudo-simulation, we use the NAVSIM
v2 one-stage NavTest evaluation protocol and report Extended Predictive
Driver Model Score (EPDMS). This evaluation is distinct from both the
NAVSIM v1/v1.1 PDMS metric and the NAVSIM v2 two-stage leaderboard
aggregation. For single-region training, OOD  EPDMS  is the unweighted mean
over the three unseen regions, whereas Total  EPDMS  is population-weighted
over all NavTest tokens. Full metric definitions are provided in
Appendix~\ref{app:navsim-metrics}.

\subsection{\textbf{Implementation Details}}

\textbf{Pretraining:}
As mentioned in the previous section, we conduct domain-specific
self-supervised pretraining on balanced NAVSIM NavTrain subsets using a
ViT-S/14 backbone for 100 epochs with AdamW and cosine learning rate
scheduling. The input resolution is $224\times224$, and the left, front, and
right camera views are used. Training is performed on 2 NVIDIA H200 GPUs with
a batch size of 64 per GPU.

\textbf{End-to-End Autonomous Driving:}
For nuScenes (LAW), models are trained for 12 epochs using AdamW with a base
learning rate of $5\times10^{-5}$, weight decay 0.01, cosine annealing
(minimum learning rate ratio $10^{-3}$), and gradient clipping with maximum
L2 norm 35. Training uses one NVIDIA L40S GPU with a batch size of 2 per GPU.
For NAVSIM (TransFuser and Latent TransFuser), models are trained for 20 epochs
using Adam with a constant learning rate of $10^{-4}$, zero weight decay, no
scheduler or warmup, and gradient clipping with maximum norm 1.0. Training
uses 4 NVIDIA H200 GPUs with a batch size of 32 per GPU (effective batch size
128). For LAW, we evaluate both frozen and fully fine-tuned navbal backbones;
the NAVSIM representation comparison in
Table~\ref{tab:navsim_representation_comparison} uses frozen pretrained
backbones. All results are reported as mean$\pm$std over three training seeds.

\subsection{\textbf{Zero-Shot Cross-City Transfer under Open-Loop Evaluation}}
We evaluate zero-shot cross-city transfer under strict city-disjoint splits on nuScenes, as summarized in Table~\ref{tab:law_zero_shot_cross_city}.

\begin{table*}[t]
    \centering
\caption{
Zero-shot cross-city transfer on nuScenes using LAW. We report joint-city,
in-domain, and cross-city evaluation under Boston--Singapore splits.
L2 (m) and collision rate (Coll, \%) are averaged over the 3\,s prediction
horizon; lower is better. Results are mean$\pm$std over three
downstream-training seeds. \textit{Swin} denotes the original LAW backbone,
\textit{Public} denotes large-scale pretrained models, and
\textit{navbal} denotes our ViT-S/14 models pretrained on balanced
NAVSIM NavTrain subsets. The final row reports the difference between the Swin baseline and the best
SSL result in each column (Swin $-$ Best SSL); positive values indicate an
improvement from SSL. 
}
    \label{tab:law_zero_shot_cross_city}
    \small
    \setlength{\tabcolsep}{5pt}
    \renewcommand{\arraystretch}{1.2}
    \resizebox{\textwidth}{!}{%
    \begin{tabular}{|c|l|cc|cc|cc|cc|cc|}
    \toprule
    \multirow{2}{*}{\textbf{Pretraining}}
    & \multirow{2}{*}{\textbf{Backbone}}
    & \multicolumn{2}{c|}{\textbf{Multi-Domain B+S $\rightarrow$ B+S}}
    & \multicolumn{2}{c|}{\textbf{In Domain B $\rightarrow$ B}}
    & \multicolumn{2}{c|}{\textbf{Cross Domain B $\rightarrow$ S}}
    & \multicolumn{2}{c|}{\textbf{In Domain S $\rightarrow$ S}}
    & \multicolumn{2}{c|}{\textbf{Cross Domain S $\rightarrow$ B}} \\
    &
    & \textbf{L2 @avg} $\downarrow$ & \textbf{Coll @avg} $\downarrow$
    & \textbf{L2 @avg} $\downarrow$ & \textbf{Coll @avg} $\downarrow$
    & \textbf{L2 @avg} $\downarrow$ & \textbf{Coll @avg} $\downarrow$
    & \textbf{L2 @avg} $\downarrow$ & \textbf{Coll @avg} $\downarrow$
    & \textbf{L2 @avg} $\downarrow$ & \textbf{Coll @avg} $\downarrow$ \\
    \midrule

    Supervised
    & Swin-Tiny (original LAW)
    & 0.72$\pm$0.05
    & 0.35$\pm$0.08
    & 0.67$\pm$0.03
    & 0.32$\pm$0.02
    & 4.06$\pm$0.51
    & 5.66$\pm$0.35
    & 0.97$\pm$0.17
    & 0.54$\pm$0.17
    & {1.11$\pm$0.19}
    & {0.84$\pm$0.52} \\

    \midrule

    \multirow{9}{*}{\shortstack[c]{Self-\\Supervised}}
    & DINOv2 (ViT-g/14, public, frozen)
    & 0.85$\pm$0.04
    & 0.31$\pm$0.02
    & 0.68$\pm$0.04
    & 0.27$\pm$0.03
    & {3.00$\pm$0.16}
    & 3.91$\pm$0.69
    & 1.37$\pm$0.04
    & 0.60$\pm$0.05
    & 2.17$\pm$0.46
    & 1.48$\pm$1.10 \\

    & I-JEPA (ViT-H/14, public, frozen)
    & 0.72$\pm$0.03
    & 0.41$\pm$0.33
    & 0.65$\pm$0.02
    & \textbf{0.26$\pm$0.04}
    & 3.53$\pm$1.23
    & {2.64$\pm$1.22}
    & 0.98$\pm$0.04
    & \textbf{0.29$\pm$0.01}
    & 1.99$\pm$0.64
    & 1.29$\pm$0.49 \\

    & MAE (ViT-H/14, public, frozen)
    & 0.67$\pm$0.01
    & 0.34$\pm$0.17
    & \textbf{0.59$\pm$0.01}
    & 0.32$\pm$0.02
    & 3.06$\pm$0.97
    & 3.92$\pm$1.52
    & 0.79$\pm$0.03
    & \textbf{0.29$\pm$0.05}
    & 3.61$\pm$1.23
    & 1.94$\pm$1.12 \\

    \cmidrule(lr){2-12}

    & DINOv2 (ViT-S/14, navbal, frozen)
    & 1.67$\pm$0.15
    & 1.06$\pm$0.48
    & 1.07$\pm$0.03
    & 0.37$\pm$0.02
    & 4.04$\pm$0.14
    & 5.12$\pm$0.36
    & 1.64$\pm$0.02
    & 1.00$\pm$0.10
    & 3.82$\pm$0.01
    & 3.53$\pm$0.12 \\

    & DINOv2 (ViT-S/14, navbal, fine-tuned)
    & 1.64$\pm$0.04
    & 1.01$\pm$0.14
    & 1.00$\pm$0.01
    & 0.37$\pm$0.02
    & 3.89$\pm$0.03
    & 4.97$\pm$0.43
    & 1.60$\pm$0.03
    & 1.14$\pm$0.06
    & 3.82$\pm$0.01
    & 2.39$\pm$0.04 \\

    & I-JEPA (ViT-S/14, navbal, frozen)
    & 0.65$\pm$0.30
    & 0.31$\pm$0.11
    & 0.61$\pm$0.21
    & 0.27$\pm$0.03
    & 1.88$\pm$0.06
    & 2.60$\pm$0.18
    & \textbf{0.71$\pm$0.01}
    & 0.33$\pm$0.03
    & 1.84$\pm$0.01
    & 1.47$\pm$0.05 \\

    & I-JEPA (ViT-S/14, navbal, fine-tuned)
    & \textbf{0.63$\pm$0.07}
    & \textbf{0.29$\pm$0.11}
    & 0.63$\pm$0.05
    & 0.31$\pm$0.09
    & \textbf{0.70$\pm$0.08}
    & \textbf{0.33$\pm$0.09}
    & 0.73$\pm$0.04
    & 0.46$\pm$0.10
    & \textbf{0.82$\pm$0.01}
    & \textbf{0.45$\pm$0.12} \\

    & MAE (ViT-S/14, navbal, frozen)
    & 1.45$\pm$0.15
    & 1.36$\pm$0.37
    & 1.25$\pm$0.09
    & 0.65$\pm$0.02
    & 3.95$\pm$0.06
    & 4.93$\pm$0.26
    & 1.45$\pm$0.11
    & 1.00$\pm$0.14
    & 3.66$\pm$0.12
    & 3.39$\pm$0.31 \\

    & MAE (ViT-S/14, navbal, fine-tuned)
    & 1.33$\pm$0.08
    & 1.27$\pm$0.18
    & 1.10$\pm$0.03
    & 0.56$\pm$0.05
    & 3.89$\pm$0.10
    & 5.00$\pm$0.17
    & 1.24$\pm$0.02
    & 1.18$\pm$0.16
    & 3.33$\pm$0.02
    & 2.92$\pm$0.09 \\

    \midrule
    \multicolumn{2}{|c|}{$\Delta$ (Swin $-$ Best SSL)}
    & +0.09
    & +0.06
    & +0.08
    & +0.06
    & +3.36
    & +5.33
    & +0.26
    & +0.25
    & +0.29
    & +0.39 \\

    \bottomrule
    \end{tabular}}
\end{table*}

\begin{table}[t]
    \centering
    \caption{
Test-time horizontal-flip diagnostic for the cross-city LAW evaluation in
Table~\ref{tab:law_zero_shot_cross_city}. At test time, camera images are
horizontally flipped before inference, and the predicted trajectories are
horizontally flipped back to the original coordinate frame before evaluation.
Metrics and all other aspects of the evaluation protocol are unchanged.
}
\label{tab:law_mirrored_cross_city}
    \small
    \setlength{\tabcolsep}{4pt}
    \renewcommand{\arraystretch}{1.2}
    \resizebox{\columnwidth}{!}{%
    \begin{tabular}{|l|cc|cc|}
    \toprule
    \multirow{2}{*}{\textbf{Backbone}}
    & \multicolumn{2}{c|}{\textbf{Flipped B $\rightarrow$ S}}
    & \multicolumn{2}{c|}{\textbf{Flipped S $\rightarrow$ B}} \\
    & \textbf{L2 @avg} $\downarrow$ & \textbf{Coll @avg} $\downarrow$ &
    \textbf{L2 @avg} $\downarrow$ & \textbf{Coll @avg} $\downarrow$ \\
    \midrule
    Swin-Tiny (supervised, original LAW)
    & 4.02$\pm$0.48 & 5.68$\pm$0.75
    & \textbf{1.15$\pm$0.19} & \textbf{0.84$\pm$0.63} \\
    \midrule
    DINOv2 (ViT-g/14, public, frozen)
        & \textbf{2.86$\pm$0.09} & 3.34$\pm$0.97
        & 2.15$\pm$0.44 & 1.75$\pm$1.31 \\
    I-JEPA (ViT-H/14, public, frozen)
        & 2.99$\pm$0.24 & \textbf{2.38$\pm$0.61}
        & 2.00$\pm$0.76 & 1.29$\pm$0.51 \\
    MAE (ViT-H/14, public, frozen)
        & 2.96$\pm$0.85 & 2.46$\pm$2.16
        & 3.61$\pm$1.25 & 1.83$\pm$1.11 \\
    \midrule
    DINOv2 (ViT-S/14, navbal, frozen)
        & 4.04$\pm$0.15 & 5.98$\pm$0.33
        & 3.82$\pm$0.04 & 6.58$\pm$0.09 \\
    I-JEPA (ViT-S/14, navbal, frozen)
        & 3.89$\pm$0.07 & 5.61$\pm$0.93
        & 1.80$\pm$0.03 & 3.80$\pm$0.04 \\
    MAE (ViT-S/14, navbal, frozen)
        & 3.95$\pm$0.06 & 6.05$\pm$0.28
        & 3.68$\pm$0.10 & 6.55$\pm$0.22 \\
    \bottomrule
    \end{tabular}}
\end{table}

\paragraph{\textbf{Multi-Domain Performance}}
Columns 2 and 3 (L2@avg and Coll@avg under B+S$\rightarrow$B+S) report multi-domain performance using the standard train/test split of nuScenes, where models are trained jointly on Boston and Singapore and evaluated on the official validation set. This setting serves as a reference in the absence of an explicit geographic domain shift.

\paragraph{\textbf{Public Pretrained Backbones}}
Cross-city transfer exposes a large degradation that is not visible
in-domain. For Swin, Boston$\rightarrow$Singapore increases L2 from
0.67 to 4.06\,m and collision rate from 0.32\% to 5.66\%.
Public SSL backbones substantially reduce this degradation:
DINOv2 lowers L2 to 3.00\,m, a 26\% reduction, while I-JEPA lowers
collision rate to 2.64\%, a 53\% reduction. The best representation differs
across metrics, showing that pretraining changes both the magnitude and the
failure profile under geographic shift. In the reverse
Singapore$\rightarrow$Boston direction, the same gains do not appear,
revealing a strongly directional effect of representation pretraining.

\paragraph{\textbf{Domain-Specific SSL Pretraining}}
Domain-specific SSL pretraining produces substantially different cross-city
transfer behavior across SSL objectives and adaptation settings. The strongest
results are obtained with the fine-tuned I-JEPA ViT-S/14 backbone. For
Boston$\rightarrow$Singapore, it reduces L2 from 4.06\,m for Swin to
0.70\,m and collision rate from 5.66\% to 0.33\%. The improvement also holds
in the reverse Singapore$\rightarrow$Boston direction, where L2 decreases
from 1.11\,m to 0.82\,m and collision rate from 0.84\% to 0.45\%.
Notably, I-JEPA maintains cross-city performance close to its corresponding
in-domain performance in both directions.

\paragraph{\textbf{Generalization Gap Analysis}}
The transfer gap is highly asymmetric across directions and representations.
Relative to Swin, the best SSL configuration reduces
Boston$\rightarrow$Singapore error by 3.36\,m in L2 and 5.33 percentage
points in collision rate, whereas the corresponding gains for
Singapore$\rightarrow$Boston are only 0.29\,m and 0.39 percentage points.
Thus, representation pretraining has a disproportionately large effect in the
Boston$\rightarrow$Singapore direction, indicating that the severity of
geographic shift depends on the interaction between the source domain and the
learned representation rather than on backbone quality alone.

The test-time horizontal-flip diagnostic provides a direct test of whether this asymmetry is primarily caused by the difference between right- and left-hand traffic. Table~\ref{tab:law_mirrored_cross_city} shows that horizontal flipping does not systematically improve transfer and substantially degrades several models. The Boston--Singapore gap therefore cannot be reduced to a simple left--right geometric mismatch.


\begin{table*}[t]
    \centering
    \caption{
    NAVSIM representation comparison with frozen pretrained backbones.
    Under single-region training, values report mean OOD  EPDMS  over the three
    unseen regions; under joint four-region training, values report Total EPDMS
    on NavTest. Results are mean$\pm$std over three downstream-training seeds;
    higher is better. Best performance in each column is shown in bold.
    }
    \label{tab:navsim_representation_comparison}
    \small
    \setlength{\tabcolsep}{3.5pt}
    \renewcommand{\arraystretch}{1.2}
    \resizebox{\textwidth}{!}{%
    \begin{tabular}{|c|l|cc|cc|cc|cc|cc|}
    \toprule
    \multirow{2}{*}{\textbf{Pretraining}}
    & \multirow{2}{*}{\textbf{Backbone}}
    & \multicolumn{2}{c|}{\textbf{Las Vegas}}
    & \multicolumn{2}{c|}{\textbf{Boston}}
    & \multicolumn{2}{c|}{\textbf{Pittsburgh}}
    & \multicolumn{2}{c|}{\textbf{Singapore}}
    & \multicolumn{2}{c|}{\textbf{4-region joint}} \\
    &
    & \textbf{TF} & \textbf{LTF}
    & \textbf{TF} & \textbf{LTF}
    & \textbf{TF} & \textbf{LTF}
    & \textbf{TF} & \textbf{LTF}
    & \textbf{TF} & \textbf{LTF} \\
    \midrule

    Supervised
    & ResNet34
    & $46.8{\pm}2.7$ & $46.5{\pm}1.6$
    & $53.1{\pm}0.5$ & $50.0{\pm}2.6$
    & $58.2{\pm}2.0$ & $56.6{\pm}1.1$
    & $\mathbf{46.7{\pm}3.7}$ & $44.5{\pm}2.9$
    & $72.8{\pm}1.6$ & $69.6{\pm}1.7$ \\

    \midrule

    \multirow{6}{*}{\shortstack[c]{Self-\\Supervised}}
    & DINOv2 (ViT-g/14, public)
    & $\mathbf{52.2{\pm}0.6}$ & $\mathbf{52.7{\pm}1.3}$
    & $52.8{\pm}4.3$ & $\mathbf{59.5{\pm}3.2}$
    & $55.9{\pm}5.6$ & $\mathbf{60.4{\pm}1.0}$
    & $43.7{\pm}5.2$ & $\mathbf{47.9{\pm}4.3}$
    & $69.9{\pm}7.2$ & $\mathbf{72.6{\pm}3.5}$ \\

    & I-JEPA (ViT-H/14, public)
    & $46.8{\pm}1.1$ & $48.4{\pm}1.9$
    & $52.1{\pm}2.1$ & $51.9{\pm}3.5$
    & $58.6{\pm}0.9$ & $59.6{\pm}1.3$
    & $40.2{\pm}4.6$ & $45.7{\pm}2.9$
    & $68.4{\pm}5.4$ & $69.4{\pm}4.2$ \\

    & MAE (ViT-H/14, public)
    & $49.4{\pm}3.2$ & $51.0{\pm}2.9$
    & $54.2{\pm}0.6$ & $52.0{\pm}6.0$
    & $57.1{\pm}6.3$ & $60.3{\pm}0.7$
    & $38.0{\pm}6.1$ & $45.7{\pm}3.3$
    & $70.5{\pm}2.6$ & $68.5{\pm}5.6$ \\

    \cmidrule(lr){2-12}

    & DINOv2 (ViT-S/14, navbal)
    & $50.0{\pm}2.2$ & $49.0{\pm}2.8$
    & $52.7{\pm}2.1$ & $51.2{\pm}3.2$
    & $58.5{\pm}2.3$ & $50.2{\pm}4.3$
    & $38.1{\pm}2.2$ & $45.0{\pm}2.8$
    & $70.6{\pm}2.3$ & $70.2{\pm}3.0$ \\

    & I-JEPA (ViT-S/14, navbal)
    & $46.5{\pm}2.2$ & $48.3{\pm}2.0$
    & $49.9{\pm}3.8$ & $49.8{\pm}2.5$
    & $53.2{\pm}1.8$ & $55.7{\pm}1.9$
    & $38.8{\pm}4.1$ & $43.6{\pm}2.3$
    & $73.5{\pm}5.1$ & $72.4{\pm}0.2$ \\

    & MAE (ViT-S/14, navbal)
    & $48.4{\pm}1.6$ & $50.2{\pm}2.1$
    & $\mathbf{54.8{\pm}1.3}$ & $52.5{\pm}0.9$
    & $\mathbf{59.1{\pm}1.9}$ & $60.2{\pm}3.7$
    & $39.5{\pm}2.2$ & $46.0{\pm}1.9$
    & $\mathbf{73.9{\pm}2.0}$ & $72.3{\pm}3.6$ \\

    \bottomrule
    \end{tabular}%
    }
\end{table*}

\subsection{\textbf{Zero-Shot Cross-City Transfer under Pseudo-Simulation Evaluation} } 
We evaluate zero-shot cross-city transfer in a pseudo-simulation using NAVSIM across two architectures:
TransFuser and Latent TransFuser. In all single-city settings,
models are trained on one city and evaluated on the remaining
cities without adaptation. We summarize the results in
Table~\ref{tab:navsim_representation_comparison}.

\paragraph{\textbf{Multi-Domain Performance}}
Under standard multi-city training, supervised ResNet34 provides a strong
baseline in both planning architectures. SSL-pretrained backbones perform
competitively in this setting. MAE navbal achieves the highest TransFuser
Total EPDMS at 73.9, while public DINOv2 achieves the highest Latent TransFuser
score at 72.6, compared with 72.8 and 69.6 for ResNet34, respectively.

\paragraph{\textbf{Public Pretrained Backbones}}
For each training city and architecture, we treat ResNet34 as the supervised
baseline. For TransFuser, ResNet34 achieves OOD EPDMS values of 46.8, 53.1,
58.2, and 46.7 when trained on Las Vegas, Boston, Pittsburgh, and Singapore,
respectively. Public SSL backbones improve over this baseline in several
settings, with DINOv2 reaching 52.2 when trained on Las Vegas and MAE reaching
54.2 when trained on Boston. The gains are more consistent in Latent
TransFuser: public DINOv2 reaches 52.7, 59.5, 60.4, and 47.9 across the four
training regions, exceeding the corresponding ResNet34 baseline in every case.
These results show that large-scale public pretraining can improve zero-shot
cross-region transfer, particularly when the planner relies entirely on the
visual representation.

\paragraph{\textbf{Domain-Specific SSL Pretraining}}
To analyze the impact of domain-specific representation learning, we evaluate
ViT-S/14 backbones pretrained on balanced NAVSIM NavTrain subsets. Improvements
are source- and architecture-dependent. In TransFuser, navbal MAE exceeds ResNet34 when trained on
Las Vegas, Boston, and Pittsburgh, reaching 48.4, 54.8, and
59.1 OOD EPDMS, respectively, while Singapore-trained SSL
models remain below the supervised baseline. In Latent TransFuser, navbal MAE reaches 50.2, 52.5, 60.2, and 46.0
across Las Vegas, Boston, Pittsburgh, and Singapore training,
respectively, giving a higher mean EPDMS than ResNet34 in all four
settings. Among the domain-specific SSL variants, MAE provides the most
consistent cross-region performance.

\paragraph{\textbf{Generalization Gap Analysis}}
Zero-shot cross-city results reveal a clear structural pattern. Under
four-region training, performance differences between backbones are relatively
small, with Total EPDMS ranging from 68.4 to 73.9 for TransFuser and from 68.5
to 72.6 for Latent TransFuser. Under strict single-region training, the
differences become substantially larger and depend strongly on both the source
region and visual backbone. The effect of representation is particularly
pronounced in Latent TransFuser, where the strongest SSL backbone exceeds
ResNet34 by 6.2, 9.5, 3.8, and 3.4 OOD EPDMS points for Las Vegas, Boston,
Pittsburgh, and Singapore training, respectively. These results show that joint four-region training reduces performance
differences across visual backbones, whereas single-region training exposes
substantially larger and source-dependent representation sensitivity. Detailed region-level results and directional EPDMS component breakdowns are
provided in Appendix~\ref{app:detailed-results}.

\begin{figure}[t]
\centering
\includegraphics[width=\columnwidth]{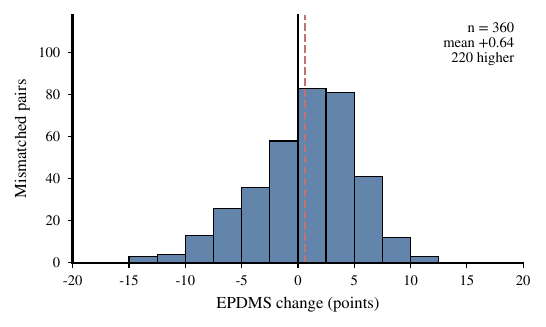}
\caption{
EPDMS change under the test-time horizontal-flip diagnostic across 360
source--target evaluations with opposite driving-side conventions.
}
\label{fig:mirror-delta-hist}
\end{figure}

To examine whether the geographic transfer differences are primarily driven by left- versus right-hand traffic, we apply the test-time horizontal-flip diagnostic to source--target pairs with opposite driving-side conventions. Across 360 evaluations, horizontal flipping increases EPDMS in 220 cases and decreases it in 140, with a mean change of only $+0.64$ points (Fig.~\ref{fig:mirror-delta-hist}). This small average change indicates that driving-side differences alone do not explain the observed geographic transfer degradation.

\begin{table}[t]
    \centering
    \caption{
Scenario composition of the balanced NAVSIM training subsets.
Vehicle statistics are computed from the current frame, while stationary,
turning, and ego-speed measures are derived from the future ego trajectory.
The ego-speed proxy is forward displacement divided by the 4\,s horizon.
}
    \label{tab:scenario-coverage}
    \footnotesize
    \setlength{\tabcolsep}{2.7pt}
    \renewcommand{\arraystretch}{1.2}
    \begin{tabular}{lrrrr}
    \toprule
     & \textbf{Boston} & \textbf{Las Vegas} & \textbf{Pitts.} & \textbf{Sing.} \\
    \midrule
    Vehicles within 20\,m       & 4.96 & 4.84 & 2.57 & 0.72 \\
    Vehicles in ego corridor    & 0.63 & 0.81 & 0.44 & 0.10 \\
    Stationary share            & 3.7\% & 4.0\% & 2.0\% & 0.5\% \\
    Turning share & 41\% & 31\% & 46\% & 67\% \\
    Ego-speed proxy (m/s)       & 4.95 & 3.98 & 5.68 & 5.37 \\
    Nearest vehicle (m)         & 11.2 & 10.6 & 27.4 & 51.3 \\
    \bottomrule
    \end{tabular}
\end{table}

\section{\textbf{DISCUSSION}}
\paragraph{\textbf{What Drives Cross-City Transfer Degradation}}
Our results show that cross-city degradation depends on transfer direction,
evaluation metric, visual representation, and planning architecture. The
degradation is strongly asymmetric across transfer directions, while the
test-time horizontal-flip diagnostic indicates that driving-side orientation
alone does not explain this asymmetry. No single SSL objective dominates
across all settings: I-JEPA is strongest in LAW, while DINOv2 and MAE are
stronger in different NAVSIM configurations, indicating that SSL gains are
setting-dependent. Moreover, L2 error and collision rate do not always agree,
indicating that trajectory proximity alone does not fully capture safety under
transfer. Representation differences are particularly pronounced in Latent
TransFuser, where the LiDAR branch is removed, while TransFuser retains the
supervised LiDAR branch.

\paragraph{\textbf{Geographic Structure and the Content of the Source Driving Scenario}}

Cross-region transfer remains strongly dependent on the source region across
both TransFuser and Latent TransFuser and across supervised and self-supervised
backbones, even after controlling for training-set size. We also find that
region identity remains highly recoverable from the pretrained visual
representations (97--99\% linear-probe accuracy;
Appendix~\ref{app:region-probe}), indicating that substantial geographic
information remains encoded after SSL pretraining.

\begin{figure}[t]
\centering
\includegraphics[width=\columnwidth]{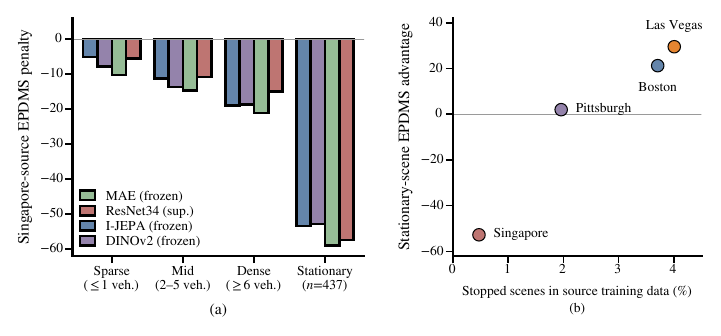}
\caption{
Source scenario content and NAVSIM transfer degradation.
(a) Singapore-source EPDMS penalty across sparse-, mid-, dense-traffic,
and stationary test scenes on Boston, Las Vegas, and Pittsburgh; the
stationary subset contains 437 samples.
(b) Relationship between the percentage of stopped scenes in the source
training data and stationary-scene performance, averaged over target
regions and backbone families.
}
\label{fig:penalty-scene}
\end{figure}

As shown in Table~\ref{tab:scenario-coverage}, the balanced NAVSIM source
regions also differ substantially in their driving-scenario composition.
Singapore contains fewer stationary ego scenarios, in which the ego vehicle
remains stopped or moves only minimally over the future horizon, and has
substantially lower surrounding-vehicle density than Boston and Las Vegas.
Fig.~\ref{fig:penalty-scene} examines this effect from two complementary
perspectives. Fig.~\ref{fig:penalty-scene}(a) shows where the
Singapore-source performance deficit occurs across scenario types:
Singapore-trained models exhibit their largest deficits on stationary and
dense-traffic scenes across the evaluated backbone families.
Fig.~\ref{fig:penalty-scene}(b) compares source stationary-scene frequency
with performance on stationary test scenes, showing that source regions with
more stopped scenes tend to achieve stronger stationary-scene performance.

Detailed scenario definitions and additional analyses are provided in
Appendix~\ref{app:source-coverage}. Together, these results show that
cross-region generalization depends not only on the visual representation,
but also on the driving scenarios represented in the source training data,
highlighting the importance of representative data curation for robust
autonomous driving.

\begin{figure}[t]
\centering
\includegraphics[width=\columnwidth]{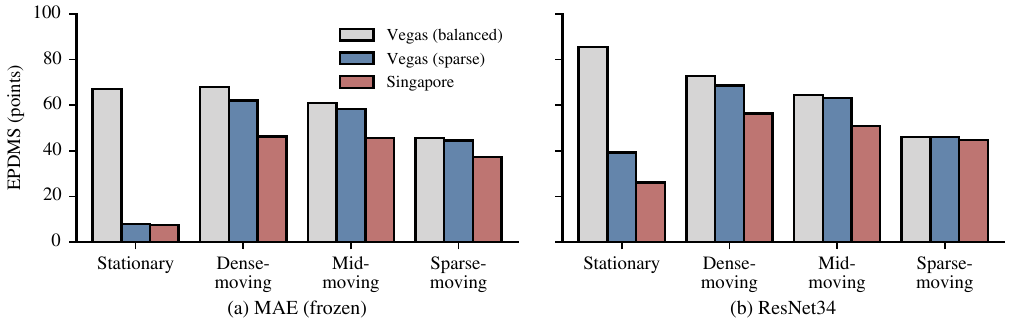}
\caption{
Source content design experiment, EPDMS results on NAVSIM. Training on the sparse Las Vegas subset produces the largest degradation on stationary
scenes, with smaller changes on moving scenes.
}
\label{fig:density-control}
\end{figure}

\paragraph{\textbf{Controlled Source-Scenario Composition}} 
We use MAE as the SSL representative because it provides the most consistent
domain-specific cross-region performance in NAVSIM (Table~\ref{tab:navsim_representation_comparison}),
while ResNet34 serves as the supervised control. We construct a sparse Las Vegas subset with the same 10{,}821 downstream
samples as the balanced training set, but containing only moving-ego scenes
with at most one nearby vehicle and no stationary examples. For MAE, the
balanced and sparse conditions use the same frozen Las Vegas-pretrained
ViT-S/14 checkpoint. Stationary-scene EPDMS decreases from 67.1 to 7.9 for
MAE, a drop of 59.2 points (95\% CI $[53.0,65.4]$), and decreases by
46.3 points for ResNet34 (95\% CI $[40.6,51.6]$). This effect remains 
when evaluation is restricted to unseen regions, but is substantially smaller on moving scenes, as shown in Fig.~\ref{fig:density-control}. Detailed stationary- and moving-scene results, including the OOD-only
analysis, are provided in Appendix~\ref{app:source-coverage}. These results show that the source training distribution directly affects
generalization to stationary scenes. Since the sparse subset changes both
the frequency of stationary examples and surrounding traffic density, the
effect reflects the overall composition of the source training data.

\begin{figure}[t]
    \centering
    \includegraphics[width=\columnwidth]
    {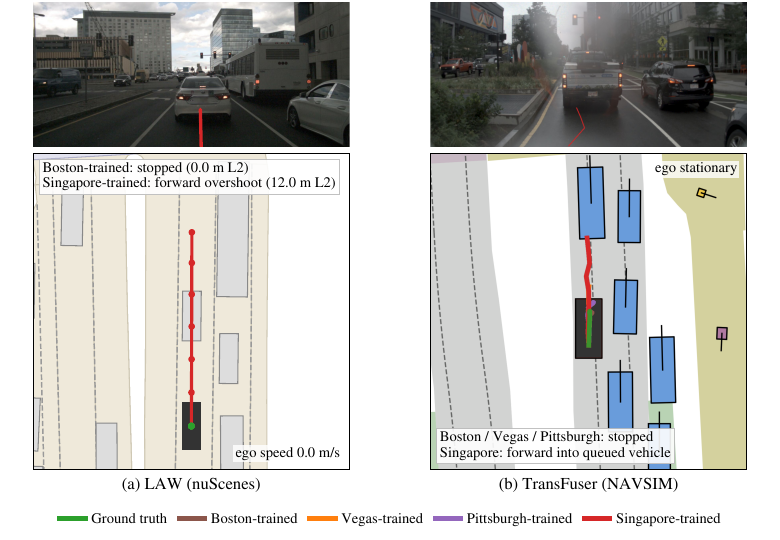}
    \caption{
Representative low-speed behavior across benchmarks using the same frozen
MAE ViT-S/14 architecture with source-specific SSL checkpoints.
(a) LAW on nuScenes. (b) TransFuser on NAVSIM.
In both examples, the Singapore-trained model predicts excessive forward
motion in a stationary scene. 
}
    \label{fig:qualitative-low-speed}
\end{figure}

\paragraph{\textbf{Cross-Benchmark Low-Speed Behavior}}
A related low-speed pattern appears in LAW/nuScenes. Using the frozen
MAE ViT-S/14 backbone with source-specific SSL checkpoints, we
define stopped samples as those with CAN-bus ego speed below
$0.3\,\mathrm{m/s}$. On stopped validation samples, LAW achieves an L2
error of 0.30\,m for Boston$\rightarrow$Boston, 6.69\,m for
Boston$\rightarrow$Singapore, 5.56\,m for Singapore$\rightarrow$Boston,
and 5.13\,m for Singapore$\rightarrow$Singapore.
Fig.~\ref{fig:qualitative-low-speed} shows the corresponding qualitative
behavior: the Boston-trained model remains close to the stationary
ground-truth trajectory, whereas the Singapore-trained model predicts
substantial forward motion. A similar low-speed failure pattern appears in
NAVSIM. The controlled NAVSIM design confirms the strong effect of the content of the source, while the nuScenes results show that the same low-speed failure
pattern appears across benchmarks.

\paragraph{\textbf{Limitations}} 
Our evaluation spans two cities in open-loop and four regions in
pseudo-simulation settings; broader geographic diversity would further
strengthen external validity. We do not explicitly disentangle weather,
lighting, or seasonal effects from geographic domain shift. The LiDAR encoder
remains a supervised ResNet34, and our study focuses exclusively on
image-based self-supervised learning. Due to the substantial computational
cost of large-scale video pretraining, we do not include V-JEPA in the current
study.
Extending the study to broader geographic coverage, video-based SSL, and
multimodal self-supervised objectives are important directions for future
work.

\begin{table*}[!t]
\centering
\caption{
EPDMS sub-component breakdown for Pittsburgh- and Singapore-trained
Latent TransFuser models. Values are mean$\pm$std over three
downstream-training seeds. Models are trained on one source region and
evaluated both in-domain (ID) and zero-shot on the remaining regions.
DINOv2 and I-JEPA use frozen public pretrained backbones; ResNet34 is the
supervised reference. Best performance for each fixed train--evaluation
pair and metric is shown in bold.
}
\label{tab:pdms-directional-component-breakdown}

\scriptsize
\setlength{\tabcolsep}{2.5pt}
\renewcommand{\arraystretch}{1.2}

\resizebox{\textwidth}{!}{%
\begin{tabular}{|lll|c|c|c|c|c|c|c|c|c|c|}
\toprule
\textbf{Backbone}
& \textbf{Train}
& \textbf{Eval region}
& \textbf{NC}
& \textbf{DAC}
& \textbf{DDC}
& \textbf{TLC}
& \textbf{EP}
& \textbf{TTC}
& \textbf{LK}
& \textbf{HC}
& \textbf{TFEC}
& \textbf{EPDMS} \\
\midrule


\multirow{8}{*}{\shortstack[l]{DINOv2\\ViT-g/14}}
& \multirow{4}{*}{Pittsburgh}
& Pittsburgh (ID)
& $\mathbf{93.6\pm1.3}$ & $75.9\pm5.8$ & $91.4\pm2.3$ & $\mathbf{99.8\pm0.2}$
& $86.1\pm1.5$ & $92.3\pm1.0$ & $84.8\pm4.1$ & $96.2\pm0.4$
& $35.2\pm5.1$ & $\mathbf{60.8\pm4.4}$ \\

&
& Las Vegas
& $\mathbf{91.1\pm1.2}$ & $\mathbf{87.6\pm2.4}$ & $98.8\pm0.4$ & $\mathbf{100.0\pm0.0}$
& $87.1\pm2.0$ & $86.9\pm1.1$ & $\mathbf{89.9\pm2.2}$ & $98.7\pm0.5$
& $42.0\pm2.0$ & $\mathbf{72.7\pm2.4}$ \\

&
& Boston
& $88.3\pm1.1$ & $77.7\pm3.0$ & $\mathbf{93.8\pm1.6}$ & $99.6\pm0.1$
& $85.4\pm1.0$ & $84.9\pm2.0$ & $87.4\pm3.1$ & $97.9\pm0.2$
& $43.9\pm3.0$ & $\mathbf{61.8\pm2.3}$ \\

&
& Singapore
& $94.8\pm0.6$ & $\mathbf{60.6\pm4.8}$ & $\mathbf{90.5\pm2.7}$ & $\mathbf{99.8\pm0.0}$
& $\mathbf{72.3\pm2.0}$ & $93.6\pm0.5$ & $78.0\pm2.6$ & $98.2\pm0.7$
& $24.2\pm7.9$ & $\mathbf{46.8\pm3.0}$ \\

\cmidrule(lr){2-13}

& \multirow{4}{*}{Singapore}
& Singapore (ID)
& $96.2\pm0.7$ & $59.6\pm4.4$ & $\mathbf{94.9\pm1.0}$ & $\mathbf{99.8\pm0.0}$
& $73.1\pm3.6$ & $94.8\pm1.4$ & $80.7\pm7.2$ & $98.7\pm0.2$
& $27.4\pm2.8$ & $47.3\pm2.0$ \\

&
& Las Vegas
& $75.7\pm5.2$ & $\mathbf{88.5\pm2.3}$ & $\mathbf{98.1\pm1.0}$ & $\mathbf{99.1\pm0.3}$
& $89.7\pm3.9$ & $71.4\pm6.0$ & $82.0\pm7.3$ & $99.0\pm0.2$
& $28.4\pm11.4$ & $\mathbf{57.9\pm4.5}$ \\

&
& Boston
& $73.2\pm4.2$ & $\mathbf{71.1\pm3.6}$ & $\mathbf{92.1\pm1.7}$ & $98.3\pm0.7$
& $86.4\pm3.9$ & $69.7\pm4.4$ & $84.1\pm2.7$ & $97.6\pm0.8$
& $28.1\pm9.9$ & $\mathbf{43.5\pm4.1}$ \\

&
& Pittsburgh
& $85.0\pm2.9$ & $\mathbf{65.8\pm8.3}$ & $80.8\pm6.8$ & $\mathbf{99.2\pm0.4}$
& $82.0\pm4.7$ & $83.2\pm3.3$ & $72.6\pm6.5$ & $95.6\pm0.9$
& $19.9\pm9.6$ & $\mathbf{42.3\pm5.8}$ \\

\midrule


\multirow{8}{*}{\shortstack[l]{I-JEPA\\ViT-H/14}}
& \multirow{4}{*}{Pittsburgh}
& Pittsburgh (ID)
& $92.2\pm1.6$ & $76.8\pm5.8$ & $88.0\pm7.3$ & $99.6\pm0.2$
& $86.9\pm2.4$ & $90.7\pm2.2$ & $80.5\pm4.9$ & $96.5\pm0.0$
& $28.9\pm4.9$ & $57.6\pm5.2$ \\

&
& Las Vegas
& $90.0\pm2.3$ & $\mathbf{87.6\pm1.8}$ & $\mathbf{99.0\pm0.4}$ & $99.7\pm0.3$
& $\mathbf{88.8\pm2.7}$ & $86.3\pm2.1$ & $89.0\pm2.4$ & $99.1\pm0.3$
& $35.2\pm8.3$ & $71.2\pm1.7$ \\

&
& Boston
& $88.0\pm3.9$ & $\mathbf{79.6\pm2.6}$ & $93.0\pm1.4$ & $\mathbf{99.7\pm0.2}$
& $\mathbf{86.5\pm2.8}$ & $85.4\pm4.6$ & $87.7\pm2.0$ & $\mathbf{98.2\pm0.1}$
& $36.2\pm4.7$ & $61.5\pm3.8$ \\

&
& Singapore
& $95.1\pm0.6$ & $58.6\pm2.8$ & $\mathbf{90.5\pm0.8}$ & $\mathbf{99.8\pm0.1}$
& $71.0\pm4.1$ & $94.0\pm0.6$ & $76.3\pm3.5$ & $98.7\pm0.1$
& $21.0\pm6.4$ & $46.0\pm1.7$ \\

\cmidrule(lr){2-13}

& \multirow{4}{*}{Singapore}
& Singapore (ID)
& $95.2\pm0.3$ & $60.1\pm7.5$ & $93.9\pm1.7$ & $\mathbf{99.8\pm0.0}$
& $\mathbf{73.8\pm0.8}$ & $94.6\pm0.5$ & $84.3\pm2.3$ & $98.8\pm0.1$
& $37.0\pm5.0$ & $48.5\pm5.3$ \\

&
& Las Vegas
& $75.0\pm0.2$ & $80.8\pm3.7$ & $96.2\pm2.0$ & $99.0\pm0.1$
& $\mathbf{93.2\pm0.9}$ & $72.8\pm0.8$ & $87.3\pm2.6$ & $98.8\pm0.6$
& $35.7\pm1.6$ & $54.6\pm1.0$ \\

&
& Boston
& $71.4\pm2.6$ & $67.9\pm6.0$ & $89.4\pm2.1$ & $98.1\pm0.3$
& $\mathbf{89.6\pm0.6}$ & $69.3\pm3.1$ & $\mathbf{86.8\pm2.1}$ & $98.0\pm0.3$
& $36.3\pm2.5$ & $41.8\pm4.1$ \\

&
& Pittsburgh
& $84.2\pm1.5$ & $61.5\pm6.5$ & $77.2\pm4.5$ & $98.8\pm0.2$
& $\mathbf{84.8\pm0.6}$ & $83.0\pm1.3$ & $\mathbf{78.8\pm2.7}$ & $\mathbf{96.4\pm0.2}$
& $28.0\pm2.9$ & $40.7\pm4.4$ \\

\midrule


\multirow{8}{*}{\shortstack[l]{ResNet34\\supervised}}
& \multirow{4}{*}{Pittsburgh}
& Pittsburgh (ID)
& $\mathbf{93.6\pm0.4}$ & $\mathbf{77.7\pm1.9}$ & $\mathbf{93.2\pm1.1}$ & $99.7\pm0.1$
& $\mathbf{87.1\pm1.2}$ & $\mathbf{92.8\pm0.5}$ & $\mathbf{85.9\pm1.1}$ & $\mathbf{96.6\pm0.0}$
& $\mathbf{57.0\pm7.5}$ & $60.6\pm1.0$ \\

&
& Las Vegas
& $90.3\pm1.1$ & $85.3\pm0.7$ & $97.7\pm0.5$ & $99.9\pm0.1$
& $86.2\pm2.1$ & $\mathbf{87.8\pm1.1}$ & $89.4\pm2.0$ & $\mathbf{99.2\pm0.2}$
& $\mathbf{60.7\pm4.2}$ & $68.1\pm0.2$ \\

&
& Boston
& $\mathbf{89.2\pm1.2}$ & $79.1\pm2.4$ & $93.2\pm0.9$ & $99.6\pm0.1$
& $84.5\pm1.8$ & $\mathbf{88.0\pm1.2}$ & $\mathbf{88.1\pm1.3}$ & $\mathbf{98.2\pm0.0}$
& $\mathbf{66.0\pm4.4}$ & $61.6\pm0.7$ \\

&
& Singapore
& $\mathbf{95.7\pm0.8}$ & $52.5\pm3.0$ & $86.7\pm2.6$ & $99.7\pm0.1$
& $71.8\pm1.7$ & $\mathbf{94.8\pm0.6}$ & $\mathbf{78.3\pm3.2}$ & $\mathbf{98.8\pm0.1}$
& $\mathbf{56.8\pm11.9}$ & $40.0\pm3.5$ \\

\cmidrule(lr){2-13}

& \multirow{4}{*}{Singapore}
& Singapore (ID)
& $\mathbf{96.9\pm0.6}$ & $\mathbf{63.0\pm3.7}$ & $94.6\pm0.2$ & $\mathbf{99.8\pm0.0}$
& $73.5\pm0.5$ & $\mathbf{96.0\pm0.7}$ & $\mathbf{86.6\pm1.3}$ & $\mathbf{98.9\pm0.0}$
& $\mathbf{49.3\pm9.0}$ & $\mathbf{48.7\pm2.7}$ \\

&
& Las Vegas
& $\mathbf{79.8\pm2.1}$ & $79.5\pm0.8$ & $95.5\pm0.8$ & $\mathbf{99.1\pm0.0}$
& $92.7\pm0.2$ & $\mathbf{77.2\pm3.0}$ & $\mathbf{91.7\pm2.6}$ & $\mathbf{99.3\pm0.0}$
& $\mathbf{49.2\pm9.0}$ & $55.8\pm2.6$ \\

&
& Boston
& $\mathbf{74.7\pm2.3}$ & $67.0\pm2.0$ & $89.1\pm1.8$ & $\mathbf{98.7\pm0.2}$
& $89.1\pm0.3$ & $\mathbf{72.0\pm3.2}$ & $\mathbf{86.8\pm2.2}$ & $\mathbf{98.1\pm0.1}$
& $\mathbf{50.8\pm7.3}$ & $40.4\pm2.9$ \\

&
& Pittsburgh
& $\mathbf{86.9\pm1.5}$ & $59.6\pm3.5$ & $\mathbf{81.0\pm3.1}$ & $\mathbf{99.2\pm0.2}$
& $82.8\pm0.4$ & $\mathbf{85.2\pm2.0}$ & $78.0\pm5.4$ & $\mathbf{96.4\pm0.1}$
& $\mathbf{41.6\pm6.3}$ & $37.4\pm3.3$ \\

\bottomrule
\end{tabular}%
}
\end{table*}

\section{\textbf{CONCLUSIONS}}

We presented a controlled study of zero-shot cross-city generalization in
end-to-end autonomous driving. Under strict geographic splits in both open-loop
(nuScenes) and pseudo-simulation (NAVSIM) settings, we show that transferring
across cities induces pronounced and asymmetric degradation. While multi-city
supervision mitigates this effect, single-city training reveals strong
sensitivity to representation initialization and the content of the source driving scenario.
Self-supervised pretraining, particularly domain-specific SSL, can improve
cross-city robustness relative to supervised ImageNet backbones, although the
gains depend on the dataset, architecture, metric, and adaptation setting.
Our controlled source-content experiment further shows that changing
source-scenario composition substantially degrades stationary-scene
generalization, establishing source-data composition as a direct factor
in transfer robustness. This result also motivates scenario-aware sampling as a practical strategy
for maintaining representative source coverage. These
results support zero-shot geographic transfer as a practical stress test for
robustness in end-to-end driving systems. Upon publication, all code used to
obtain the results in this paper will be released publicly on GitHub, along
with instructions for reproducing our experiments.

\appendices
\section{\textbf{Additional Analyses}}

\subsection{\textbf{NAVSIM v2 Evaluation Metric}}
\label{app:navsim-metrics}

All NAVSIM experiments use the official NAVSIM v2 one-stage NavTest
evaluation protocol with human-behavior filtering enabled. For each token,
the Extended Predictive Driver Model Score (EPDMS) is computed from
no-at-fault collision (NC), drivable-area compliance (DAC),
driving-direction compliance (DDC), traffic-light compliance (TLC),
ego progress (EP), time-to-collision (TTC), lane keeping (LK),
history comfort (HC), and extended comfort (EC):

\begin{equation}
\begin{aligned}
\mathrm{EPDMS}
&=
(\mathrm{NC}\cdot\mathrm{DAC}\cdot\mathrm{DDC}\cdot\mathrm{TLC})
\\
&\quad\times
\frac{
5\,\mathrm{EP}
+5\,\mathrm{TTC}
+2\,\mathrm{LK}
+2\,\mathrm{HC}
+2\,\mathrm{EC}
}{16}.
\end{aligned}
\label{eq:epdms}
\end{equation}



\subsection{\textbf{Source Content Analysis}}
\label{app:source-coverage}

We characterize the balanced NAVSIM training subsets using cached downstream
targets. Vehicles are counted within 20\,m of the ego in the current frame,
excluding pedestrians and cyclists. The ego corridor is defined as
$0<x<30$\,m and $|y|<3$\,m. From the 4\,s future ego trajectory, a
sample is classified as stationary when longitudinal displacement is below
2\,m and as turning when lateral displacement exceeds 2\,m or heading
change exceeds 0.3\,rad. Moving samples are grouped as sparse, mid-density,
and dense using $\leq1$, 2--5, and $\geq6$ nearby vehicles, respectively.
Despite equalizing downstream training to 10,821 samples per region, the
scenario composition differs substantially
(Table~\ref{tab:scenario-coverage}).
The full stationary NavTest set contains 444 samples. The Singapore-source
analysis excludes Singapore targets and uses 437 samples, while the OOD-only
Las Vegas analysis additionally excludes Las Vegas and uses 281 samples from
45 logs. Uncertainty is estimated with 4,000 log-clustered bootstrap
resamples over the 83 NavTest logs containing stationary samples, after
averaging token-level EPDMS across both planner architectures and three seeds.

To test source-content effects directly, we construct a matched-count sparse
Las Vegas subset of 10,821 moving-ego samples with at most one vehicle within
20\,m. Moving ego requires at least 2\,m longitudinal displacement over
4\,s. From 12,477 eligible samples, all 4,747 zero-vehicle samples and
6,074 randomly selected one-vehicle samples are retained, yielding 0.56
nearby vehicles on average and no stationary examples.
For MAE, the balanced and sparse Las Vegas conditions use the same frozen
Las Vegas-pretrained ViT-S/14 checkpoint; only the downstream training
distribution changes. ResNet34 is trained end-to-end separately under each
condition, while training-set size, downstream protocol, and evaluation data
remain unchanged.
Stationary EPDMS decreases from 67.1 to 7.9 for MAE, a 59.2-point reduction
(95\% CI $[53.0,65.4]$), and by 46.3 points for ResNet34
(95\% CI $[40.6,51.6]$). Restricting evaluation to unseen regions yields
similar reductions of 52.5 points for MAE
(95\% CI $[44.9,60.7]$) and 54.7 points for ResNet34
(95\% CI $[48.7,60.5]$). The decrease is observed across all three downstream-training seeds for both backbones.
The effect is substantially smaller on moving samples. Since the sparse
subset changes both the frequency of stationary examples and surrounding
traffic density, the effect reflects the overall content of the source
training data.


\begin{table*}[t]
    \centering
    \caption{
Detailed NAVSIM cross-region EPDMS (\%) for TransFuser and Latent
TransFuser. Models are trained either jointly on all four regions or on one
source region and evaluated on each NavTest region. For single-region
training, OOD is the unweighted mean EPDMS over the three unseen regions,
whereas Total is the population-weighted EPDMS over the full NavTest split.
OOD is not defined for four-region training and is shown as ``--''. For
single-region training, bold values denote in-domain performance; for
four-region training, bold denotes Total EPDMS. Higher is better.
}
    \label{tab:navsim_unified_cross_city_pdms}
    \tiny
    \setlength{\tabcolsep}{2.8pt}
    \renewcommand{\arraystretch}{0.70}
    \resizebox{\textwidth}{!}{%
    \begin{tabular}{|l|c|c|cccccc|cccccc|}
    \toprule
    \multirow{2}{*}{\textbf{Backbone}}
    & \multirow{2}{*}{\textbf{Adaptation}}
    & \multirow{2}{*}{\textbf{Train}}
    & \multicolumn{6}{c|}{\textbf{TransFuser EPDMS (\%)}}
    & \multicolumn{6}{c}{\textbf{Latent TransFuser EPDMS (\%)}} \\
    & &
    & \textbf{LV} & \textbf{B} & \textbf{P} & \textbf{S}
    & \textbf{OOD} & \textbf{Total}
    & \textbf{LV} & \textbf{B} & \textbf{P} & \textbf{S}
    & \textbf{OOD} & \textbf{Total} \\
    \midrule

\multirow{5}{*}{ResNet34}
& \multirow{5}{*}{Supervised}
& 4-region
& 82.5 & 74.6 & 68.1 & 54.7 & -- & \textbf{72.8}
& 79.4 & 70.2 & 62.1 & 57.1 & -- & \textbf{69.6} \\

& & LV
& \textbf{79.2} & 59.0 & 44.1 & 37.4 & 46.8 & 59.4
& \textbf{78.6} & 58.8 & 43.5 & 37.2 & 46.5 & 59.0 \\

& & B
& 70.0 & \textbf{62.7} & 43.5 & 45.8 & 53.1 & 58.6
& 65.6 & \textbf{59.2} & 40.5 & 43.8 & 50.0 & 55.2 \\

& & P
& 70.7 & 62.7 & \textbf{61.1} & 41.2 & 58.2 & 61.6
& 68.1 & 61.6 & \textbf{60.6} & 40.0 & 56.6 & 60.2 \\

& & S
& 57.9 & 42.8 & 39.3 & \textbf{50.0} & 46.7 & 48.3
& 55.8 & 40.4 & 37.4 & \textbf{48.7} & 44.5 & 46.2 \\
\midrule

\multirow{5}{*}{I-JEPA ViT-H/14}
& \multirow{5}{*}{Frozen}
& 4-region
& 78.4 & 68.2 & 59.4 & 58.7 & -- & \textbf{68.4}
& 79.1 & 71.0 & 63.7 & 53.0 & -- & \textbf{69.4} \\

& & LV
& \textbf{77.4} & 51.6 & 43.3 & 45.6 & 46.8 & 57.6
& \textbf{78.1} & 57.9 & 42.5 & 44.9 & 48.4 & 59.5 \\

& & B
& 67.7 & \textbf{58.8} & 46.1 & 42.4 & 52.1 & 56.7
& 67.1 & \textbf{58.4} & 42.0 & 46.5 & 51.9 & 56.2 \\

& & P
& 70.5 & 57.2 & \textbf{56.8} & 48.1 & 58.6 & 60.2
& 71.2 & 61.5 & \textbf{57.6} & 46.0 & 59.6 & 61.5 \\

& & S
& 49.4 & 35.3 & 35.9 & \textbf{47.7} & 40.2 & 42.1
& 54.6 & 41.8 & 40.7 & \textbf{48.5} & 45.7 & 46.9 \\

\midrule

\multirow{5}{*}{MAE ViT-H/14}
& \multirow{5}{*}{Frozen}
& 4-region
& 79.3 & 71.1 & 62.0 & 61.4 & -- & \textbf{70.5}
& 76.6 & 67.5 & 58.6 & 65.9 & -- & \textbf{68.5} \\

& & LV
& \textbf{75.4} & 54.9 & 44.0 & 49.3 & 49.4 & 58.7
& \textbf{77.6} & 57.6 & 43.6 & 51.7 & 51.0 & 60.5 \\

& & B
& 70.5 & \textbf{64.7} & 47.0 & 45.0 & 54.2 & 60.0
& 66.7 & \textbf{56.1} & 43.4 & 45.9 & 52.0 & 55.5 \\

& & P
& 70.2 & 59.2 & \textbf{55.9} & 41.8 & 57.1 & 59.5
& 71.7 & 62.5 & \textbf{56.4} & 46.6 & 60.3 & 61.8 \\

& & S
& 49.3 & 32.1 & 32.5 & \textbf{49.2} & 38.0 & 40.6
& 54.8 & 42.3 & 40.0 & \textbf{45.5} & 45.7 & 46.5 \\

\midrule

\multirow{5}{*}{DINOv2 ViT-g/14}
& \multirow{5}{*}{Frozen}
& 4-region
& 80.7 & 70.8 & 61.9 & 55.6 & -- & \textbf{69.9}
& 82.4 & 73.4 & 63.7 & 61.5 & -- & \textbf{72.6} \\

& & LV
& \textbf{79.9} & 61.1 & 50.6 & 44.9 & 52.2 & 62.7
& \textbf{78.0} & 62.6 & 51.4 & 44.0 & 52.7 & 62.6 \\

& & B
& 69.6 & \textbf{66.2} & 48.3 & 40.4 & 52.8 & 59.7
& 73.3 & \textbf{70.5} & 53.3 & 51.9 & 59.5 & 65.0 \\

& & P
& 67.9 & 55.4 & \textbf{49.0} & 44.4 & 55.9 & 56.6
& 72.7 & 61.8 & \textbf{60.8} & 46.8 & 60.4 & 62.9 \\

& & S
& 54.4 & 38.5 & 38.2 & \textbf{47.1} & 43.7 & 45.1
& 57.9 & 43.5 & 42.3 & \textbf{47.3} & 47.9 & 48.7 \\
\midrule


\multirow{5}{*}{I-JEPA ViT-S/14}
& \multirow{5}{*}{Frozen}
& 4-region
& 83.2 & 72.9 & 65.2 & 64.4 & -- & \textbf{73.5}
& 82.7 & 72.3 & 64.4 & 61.1 & -- & \textbf{72.4} \\

& & LV
& \textbf{72.8} & 52.5 & 42.1 & 44.8 & 46.5 & 56.0
& \textbf{77.9} & 58.2 & 43.8 & 42.8 & 48.3 & 59.5 \\

& & B
& 63.8 & \textbf{57.8} & 42.6 & 43.4 & 49.9 & 54.5
& 62.7 & \textbf{61.9} & 42.7 & 43.9 & 49.8 & 55.5 \\

& & P
& 65.3 & 51.7 & \textbf{47.9} & 42.7 & 53.2 & 54.1
& 67.8 & 56.3 & \textbf{51.8} & 43.1 & 55.7 & 57.2 \\

& & S
& 46.6 & 35.3 & 34.6 & \textbf{35.5} & 38.8 & 39.0
& 54.5 & 38.6 & 37.7 & \textbf{45.3} & 43.6 & 44.8 \\

\midrule

\multirow{5}{*}{I-JEPA ViT-S/14}
& \multirow{5}{*}{Fine-tuned}
& 4-region
& 80.0 & 69.1 & 58.3 & 68.5 & -- & \textbf{70.5}
& 81.5 & 72.2 & 61.4 & 70.4 & -- & \textbf{72.9} \\

& & LV
& \textbf{75.6} & 50.5 & 37.3 & 30.3 & 39.4 & 53.0
& \textbf{78.5} & 52.4 & 40.2 & 34.3 & 42.3 & 55.8 \\

& & B
& 50.4 & \textbf{48.8} & 35.5 & 32.6 & 39.5 & 44.1
& 55.5 & \textbf{58.0} & 41.4 & 35.7 & 44.2 & 50.3 \\

& & P
& 57.3 & 47.5 & \textbf{48.4} & 32.8 & 45.9 & 48.6
& 54.2 & 44.5 & \textbf{43.5} & 32.0 & 43.6 & 45.6 \\

& & S
& 41.0 & 28.7 & 33.3 & \textbf{41.0} & 34.3 & 35.7
& 44.1 & 33.8 & 34.4 & \textbf{43.4} & 37.4 & 38.9 \\

\midrule

\multirow{5}{*}{MAE ViT-S/14}
& \multirow{5}{*}{Frozen}
& 4-region
& 82.4 & 73.3 & 65.6 & 67.5 & -- & \textbf{73.9}
& 80.9 & 72.3 & 64.2 & 64.2 & -- & \textbf{72.3} \\

& & LV
& \textbf{78.5} & 55.9 & 45.6 & 43.8 & 48.4 & 59.5
& \textbf{76.7} & 57.8 & 45.9 & 47.0 & 50.2 & 60.0 \\

& & B
& 71.8 & \textbf{66.4} & 48.3 & 44.2 & 54.8 & 61.1
& 68.5 & \textbf{62.6} & 44.2 & 44.8 & 52.5 & 58.1 \\

& & P
& 72.0 & 61.8 & \textbf{59.7} & 43.6 & 59.1 & 61.9
& 71.5 & 64.0 & \textbf{60.7} & 45.2 & 60.2 & 62.9 \\

& & S
& 50.1 & 34.1 & 34.2 & \textbf{45.2} & 39.5 & 41.2
& 54.3 & 42.2 & 41.4 & \textbf{41.4} & 46.0 & 46.0 \\

\midrule

\multirow{5}{*}{MAE ViT-S/14}
& \multirow{5}{*}{Fine-tuned}
& 4-region
& 80.8 & 71.5 & 61.9 & 64.8 & -- & \textbf{71.6}
& 80.7 & 70.7 & 63.5 & 65.8 & -- & \textbf{71.8} \\

& & LV
& \textbf{80.0} & 59.3 & 48.5 & 44.3 & 50.7 & 61.7
& \textbf{79.8} & 56.0 & 44.4 & 45.2 & 48.5 & 59.9 \\

& & B
& 70.6 & \textbf{67.1} & 50.1 & 44.0 & 54.9 & 61.2
& 65.2 & \textbf{58.5} & 43.6 & 46.8 & 51.9 & 55.9 \\

& & P
& 66.2 & 55.6 & \textbf{53.0} & 38.1 & 53.3 & 55.9
& 65.3 & 56.0 & \textbf{51.2} & 40.0 & 53.8 & 55.6 \\

& & S
& 52.1 & 39.3 & 40.7 & \textbf{48.3} & 44.0 & 45.3
& 49.2 & 40.7 & 41.3 & \textbf{41.0} & 43.7 & 43.7 \\

\midrule

\multirow{5}{*}{DINOv2 ViT-S/14}
& \multirow{5}{*}{Frozen}
& 4-region
& 80.2 & 70.2 & 62.4 & 61.6 & -- & \textbf{70.6}
& 80.1 & 70.0 & 60.9 & 61.4 & -- & \textbf{70.2} \\

& & LV
& \textbf{77.1} & 56.8 & 47.0 & 46.2 & 50.0 & 60.0
& \textbf{75.4} & 55.1 & 45.8 & 46.1 & 49.0 & 58.6 \\

& & B
& 66.4 & \textbf{63.0} & 46.7 & 45.0 & 52.7 & 58.0
& 64.5 & \textbf{62.1} & 46.5 & 42.7 & 51.2 & 56.7 \\

& & P
& 71.1 & 60.3 & \textbf{58.9} & 44.2 & 58.5 & 61.1
& 60.6 & 52.0 & \textbf{47.5} & 38.1 & 50.2 & 51.8 \\

& & S
& 45.2 & 33.4 & 35.8 & \textbf{41.0} & 38.1 & 39.0
& 54.5 & 40.8 & 39.8 & \textbf{43.8} & 45.0 & 45.6 \\

\midrule

\multirow{5}{*}{DINOv2 ViT-S/14}
& \multirow{5}{*}{Fine-tuned}
& 4-region
& 75.6 & 65.4 & 56.8 & 59.0 & -- & \textbf{66.1}
& 81.4 & 71.6 & 61.6 & 67.5 & -- & \textbf{72.2} \\

& & LV
& \textbf{75.3} & 52.5 & 42.4 & 42.6 & 45.8 & 56.5
& \textbf{77.1} & 56.1 & 45.5 & 42.2 & 47.9 & 58.8 \\

& & B
& 61.6 & \textbf{54.6} & 39.1 & 44.0 & 48.2 & 52.1
& 63.9 & \textbf{62.3} & 44.4 & 40.6 & 49.6 & 55.8 \\

& & P
& 65.0 & 53.2 & \textbf{51.2} & 38.7 & 52.3 & 54.5
& 62.6 & 59.4 & \textbf{51.1} & 38.0 & 53.3 & 55.4 \\

& & S
& 48.5 & 37.2 & 37.3 & \textbf{38.9} & 41.0 & 41.3
& 57.0 & 45.7 & 42.3 & \textbf{43.5} & 48.3 & 48.4 \\

\bottomrule
    \end{tabular}}
\end{table*}

\subsection{\textbf{Geographic Information in Visual Representations}}
\label{app:region-probe}

We evaluate geographic information using a four-way linear classifier on
spatially pooled backbone features. I-JEPA, DINOv2, and MAE use the
four-region NavTrain-balanced ViT-S/14 SSL checkpoints and are evaluated before downstream planner training; ResNet34 serves as
a supervised reference. Probe training and evaluation use balanced NavTest
samples with disjoint driving logs, while the encoders remain fixed. Accuracy
is 0.99 for I-JEPA, 0.98 for DINOv2, 0.97 for MAE, and 0.87 for ResNet34,
compared with a four-way chance accuracy of 0.25. Therefore, geographic
identity remains strongly recoverable from both supervised and self-supervised
pretrained representations.

\subsection{\textbf{Directional EPDMS Component Analysis}}
\label{app:component-analysis}

Table~\ref{tab:pdms-directional-component-breakdown} shows that transfer
directions fail through different EPDMS components. For Pittsburgh$\rightarrow$Singapore, the dominant reductions occur in
DAC and EP, with additional degradation in LK and, for the SSL backbones,
TFEC, whereas Singapore$\rightarrow$Pittsburgh shows stronger reductions
in NC and TTC.

\subsection{\textbf{Detailed Cross-Region Results}}
\label{app:detailed-results}

Table~\ref{tab:navsim_unified_cross_city_pdms} reports the complete
region-level EPDMS matrix underlying the aggregate comparisons in
Table~\ref{tab:navsim_representation_comparison}.


\FloatBarrier
\clearpage

\bstctlcite{BSTcontrol}
\bibliographystyle{IEEEtran}
\bibliography{ref}

@IEEEtranBSTCTL{BSTcontrol,
  CTLuse_url       = "no",
  CTLuse_doi       = "yes",
  CTLuse_eprint    = "yes"
}

@inproceedings{assran2023selfsupervisedlearningimagesjointembedding,
  title={Self-supervised learning from images with a joint-embedding predictive architecture},
  author={Assran, Mahmoud and Duval, Quentin and Misra, Ishan and Bojanowski, Piotr and Vincent, Pascal and Rabbat, Michael and LeCun, Yann and Ballas, Nicolas},
  booktitle={Proceedings of the IEEE/CVF conference on computer vision and pattern recognition},
  pages={15619--15629},
  year={2023}
}

@article{assran2025vjepa2,
  title={V-JEPA~2: Self-Supervised Video Models Enable Understanding, Prediction and Planning},
  author={Assran, Mahmoud and Bardes, Adrien and Fan, David and Garrido, Quentin and Howes, Russell and
Komeili, Mojtaba and Muckley, Matthew and Rizvi, Ammar and Roberts, Claire and Sinha, Koustuv and Zholus, Artem and
Arnaud, Sergio and Gejji, Abha and Martin, Ada and Robert Hogan, Francois and Dugas, Daniel and
Bojanowski, Piotr and Khalidov, Vasil and Labatut, Patrick and Massa, Francisco and Szafraniec, Marc and
Krishnakumar, Kapil and Li, Yong and Ma, Xiaodong and Chandar, Sarath and Meier, Franziska and LeCun, Yann and
Rabbat, Michael and Ballas, Nicolas},
  journal={arXiv preprint arXiv:2506.09985},
  year={2025}
}

@article{bardes2024revisiting,
  title={Revisiting Feature Prediction for Learning Visual Representations from Video},
  author={Bardes, Adrien and Garrido, Quentin and Ponce, Jean and Chen, Xinlei and Rabbat, Michael and LeCun, Yann and Assran, Mahmoud and Ballas, Nicolas},
  journal={Transactions on Machine Learning Research},
  issn={2835-8856},
  year={2024},
}

@article{zhu2025selfsupervisedrepresentationlearningjoint,
  title={Self-Supervised Representation Learning with Joint Embedding Predictive Architecture for Automotive LiDAR Object Detection},
  author={Zhu, Haoran and Dong, Zhenyuan and Topollai, Kristi and Sha, Beiyao and Choromanska, Anna},
  journal={arXiv preprint arXiv:2501.04969},
  year={2025}
}

@article{oquab2024dinov2learningrobustvisual,
  title={Dinov2: Learning robust visual features without supervision},
  author={Oquab, Maxime and Darcet, Timoth{\'e}e and Moutakanni, Th{\'e}o and Vo, Huy and Szafraniec, Marc and Khalidov, Vasil and Fernandez, Pierre and Haziza, Daniel and Massa, Francisco and El-Nouby, Alaaeldin and others},
  journal={arXiv preprint arXiv:2304.07193},
  year={2023}
}

@article{russakovsky2015imagenetlargescalevisual,
  title={Imagenet large scale visual recognition challenge},
  author={Russakovsky, Olga and Deng, Jia and Su, Hao and Krause, Jonathan and Satheesh, Sanjeev and Ma, Sean and Huang, Zhiheng and Karpathy, Andrej and Khosla, Aditya and Bernstein, Michael and others},
  journal={International journal of computer vision},
  volume={115},
  number={3},
  pages={211--252},
  year={2015},
  publisher={Springer}
}

@inproceedings{he2021maskedautoencodersscalablevision,
  title={Masked autoencoders are scalable vision learners},
  author={He, Kaiming and Chen, Xinlei and Xie, Saining and Li, Yanghao and Doll{\'a}r, Piotr and Girshick, Ross},
  booktitle={Proceedings of the IEEE/CVF conference on computer vision and pattern recognition},
  pages={16000--16009},
  year={2022}
}

@article{li2024enhancing,
  title={Enhancing end-to-end autonomous driving with latent world model},
  author={Li, Yingyan and Fan, Lue and He, Jiawei and Wang, Yuqi and Chen, Yuntao and Zhang, Zhaoxiang and Tan, Tieniu},
  journal={arXiv preprint arXiv:2406.08481},
  year={2024}
}

@article{li2025navigationguidedsparsescenerepresentation,
  title={Navigation-guided sparse scene representation for end-to-end autonomous driving},
  author={Li, Peidong and Cui, Dixiao},
  journal={arXiv preprint arXiv:2409.18341},
  year={2024}
}

@inproceedings{li2025endtoenddrivingonlinetrajectory,
  title={End-to-end driving with online trajectory evaluation via bev world model},
  author={Li, Yingyan and Wang, Yuqi and Liu, Yang and He, Jiawei and Fan, Lue and Zhang, Zhaoxiang},
  booktitle={Proceedings of the IEEE/CVF International Conference on Computer Vision},
  pages={27137--27146},
  year={2025}
}

@inproceedings{zheng2025world4driveendtoendautonomousdriving,
  title={World4drive: End-to-end autonomous driving via intention-aware physical latent world model},
  author={Zheng, Yupeng and Yang, Pengxuan and Xing, Zebin and Zhang, Qichao and Zheng, Yuhang and Gao, Yinfeng and Li, Pengfei and Zhang, Teng and Xia, Zhongpu and Jia, Peng and others},
  booktitle={Proceedings of the IEEE/CVF International Conference on Computer Vision},
  pages={28632--28642},
  year={2025}
}

@article{chitta2022transfuserimitationtransformerbasedsensor,
  title={Transfuser: Imitation with transformer-based sensor fusion for autonomous driving},
  author={Chitta, Kashyap and Prakash, Aditya and Jaeger, Bernhard and Yu, Zehao and Renz, Katrin and Geiger, Andreas},
  journal={IEEE transactions on pattern analysis and machine intelligence},
  volume={45},
  number={11},
  pages={12878--12895},
  year={2022},
  publisher={IEEE}
}

@article{li2024hydramdpendtoendmultimodalplanning,
  title={Hydra-mdp: End-to-end multimodal planning with multi-target hydra-distillation},
  author={Li, Zhenxin and Li, Kailin and Wang, Shihao and Lan, Shiyi and Yu, Zhiding and Ji, Yishen and Li, Zhiqi and Zhu, Ziyue and Kautz, Jan and Wu, Zuxuan and others},
  journal={arXiv preprint arXiv:2406.06978},
  year={2024}
}

@article{yasarla2025rocarobustcrossdomainendtoend,
  title={Roca: Robust cross-domain end-to-end autonomous driving},
  author={Yasarla, Rajeev and Han, Shizhong and Cheng, Hsin-Pai and Liu, Litian and Mahajan, Shweta and Bhattacharyya, Apratim and Shi, Yunxiao and Garrepalli, Risheek and Cai, Hong and Porikli, Fatih},
  journal={arXiv preprint arXiv:2506.10145},
  year={2025}
}

@inproceedings{jiang2023vad,
  title={Vad: Vectorized scene representation for efficient autonomous driving},
  author={Jiang, Bo and Chen, Shaoyu and Xu, Qing and Liao, Bencheng and Chen, Jiajie and Zhou, Helong and Zhang, Qian and Liu, Wenyu and Huang, Chang and Wang, Xinggang},
  booktitle={Proceedings of the IEEE/CVF International Conference on Computer Vision},
  pages={8340--8350},
  year={2023}
}

@inproceedings{hu2023planningorientedautonomousdriving,
  title={Planning-oriented autonomous driving},
  author={Hu, Yihan and Yang, Jiazhi and Chen, Li and Li, Keyu and Sima, Chonghao and Zhu, Xizhou and Chai, Siqi and Du, Senyao and Lin, Tianwei and Wang, Wenhai and others},
  booktitle={Proceedings of the IEEE/CVF conference on computer vision and pattern recognition},
  pages={17853--17862},
  year={2023}
}

@article{zhang2025futureawareendtoenddrivingbidirectional,
  title={Future-aware end-to-end driving: Bidirectional modeling of trajectory planning and scene evolution},
  author={Zhang, Bozhou and Song, Nan and Li, Jingyu and Zhu, Xiatian and Deng, Jiankang and Zhang, Li},
  journal={arXiv preprint arXiv:2510.11092},
  year={2025}
}

@inproceedings{caesar2020nuscenesmultimodaldatasetautonomous,
  title={nuscenes: A multimodal dataset for autonomous driving},
  author={Caesar, Holger and Bankiti, Varun and Lang, Alex H and Vora, Sourabh and Liong, Venice Erin and Xu, Qiang and Krishnan, Anush and Pan, Yu and Baldan, Giancarlo and Beijbom, Oscar},
  booktitle={Proceedings of the IEEE/CVF conference on computer vision and pattern recognition},
  pages={11621--11631},
  year={2020}
}

@article{cao2025pseudosimulationautonomousdriving,
  title={Pseudo-simulation for autonomous driving},
  author={Cao, Wei and Hallgarten, Marcel and Li, Tianyu and Dauner, Daniel and Gu, Xunjiang and Wang, Caojun and Miron, Yakov and Aiello, Marco and Li, Hongyang and Gilitschenski, Igor and others},
  journal={arXiv preprint arXiv:2506.04218},
  year={2025}
}

@article{singh2025mattersrepresentationalignmentglobal,
  title={What matters for Representation Alignment: Global Information or Spatial Structure?},
  author={Singh, Jaskirat and Leng, Xingjian and Wu, Zongze and Zheng, Liang and Zhang, Richard and Shechtman, Eli and Xie, Saining},
  journal={arXiv preprint arXiv:2512.10794},
  year={2025}
}

@inproceedings{he2015deepresiduallearningimage,
  title={Deep residual learning for image recognition},
  author={He, Kaiming and Zhang, Xiangyu and Ren, Shaoqing and Sun, Jian},
  booktitle={Proceedings of the IEEE conference on computer vision and pattern recognition},
  pages={770--778},
  year={2016}
}

@article{dosovitskiy2021imageworth16x16words,
  title={An image is worth 16x16 words: Transformers for image recognition at scale},
  author={Dosovitskiy, Alexey and Beyer, Lucas and Kolesnikov, Alexander and Weissenborn, Dirk and Zhai, Xiaohua and Unterthiner, Thomas and Dehghani, Mostafa and Minderer, Matthias and Heigold, Georg and Gelly, Sylvain and others},
  journal={arXiv preprint arXiv:2010.11929},
  year={2020}
}

@inproceedings{liu2021swintransformerhierarchicalvision,
  title={Swin transformer: Hierarchical vision transformer using shifted windows},
  author={Liu, Ze and Lin, Yutong and Cao, Yue and Hu, Han and Wei, Yixuan and Zhang, Zheng and Lin, Stephen and Guo, Baining},
  booktitle={Proceedings of the IEEE/CVF international conference on computer vision},
  pages={10012--10022},
  year={2021}
}

@article{li2022bevformerlearningbirdseyeviewrepresentation,
  author={Li, Zhiqi and Wang, Wenhai and Li, Hongyang and Xie, Enze and Sima, Chonghao and Lu, Tong and Yu, Qiao and Dai, Jifeng},
  journal={IEEE Transactions on Pattern Analysis and Machine Intelligence}, 
  title={BEVFormer: Learning Bird’s-Eye-View Representation From LiDAR-Camera via Spatiotemporal Transformers}, 
  year={2025},
  volume={47},
  number={3},
  pages={2020-2036},
  doi={10.1109/TPAMI.2024.3515454}}

@article{wang2026drivejepavideojepameets,
  title={Drive-JEPA: Video JEPA Meets Multimodal Trajectory Distillation for End-to-End Driving},
  author={Wang, Linhan and Yang, Zichong and Bai, Chen and Zhang, Guoxiang and Liu, Xiaotong and Zheng, Xiaoyin and Long, Xiao-Xiao and Lu, Chang-Tien and Lu, Cheng},
  journal={arXiv preprint arXiv:2601.22032},
  year={2026}
}

@article{bojarski2016endendlearningselfdriving,
  title={End to end learning for self-driving cars},
  author={Bojarski, Mariusz and Del Testa, Davide and Dworakowski, Daniel and Firner, Bernhard and Flepp, Beat and Goyal, Prasoon and Jackel, Lawrence D and Monfort, Mathew and Muller, Urs and Zhang, Jiakai and others},
  journal={arXiv preprint arXiv:1604.07316},
  year={2016}
}

@article{dauner2024navsimdatadrivennonreactiveautonomous,
  title={Navsim: Data-driven non-reactive autonomous vehicle simulation and benchmarking},
  author={Dauner, Daniel and Hallgarten, Marcel and Li, Tianyu and Weng, Xinshuo and Huang, Zhiyu and Yang, Zetong and Li, Hongyang and Gilitschenski, Igor and Ivanovic, Boris and Pavone, Marco and others},
  journal={Advances in Neural Information Processing Systems},
  volume={37},
  pages={28706--28719},
  year={2024}
}

@article{guo2025ipaditerativeproposalcentricendtoend,
  title={ipad: Iterative proposal-centric end-to-end autonomous driving},
  author={Guo, Ke and Liu, Haochen and Wu, Xiaojun and Pan, Jia and Lv, Chen},
  journal={arXiv preprint arXiv:2505.15111},
  year={2025}
}

@inproceedings{sun2024sparsedriveendtoendautonomousdriving,
  title={Sparsedrive: End-to-end autonomous driving via sparse scene representation},
  author={Sun, Wenchao and Lin, Xuewu and Shi, Yining and Zhang, Chuang and Wu, Haoran and Zheng, Sifa},
  booktitle={2025 IEEE International Conference on Robotics and Automation (ICRA)},
  pages={8795--8801},
  year={2025},
  organization={IEEE}
}

@article{chen2024end,
  title={End-to-end autonomous driving: Challenges and frontiers},
  author={Chen, Li and Wu, Penghao and Chitta, Kashyap and Jaeger, Bernhard and Geiger, Andreas and Li, Hongyang},
  journal={IEEE Transactions on Pattern Analysis and Machine Intelligence},
  volume={46},
  number={12},
  pages={10164--10183},
  year={2024},
  publisher={IEEE}
}

@article{dong2024generalizingendtoendautonomousdriving,
  title={Generalizing end-to-end autonomous driving in real-world environments using zero-shot llms},
  author={Dong, Zeyu and Zhu, Yimin and Li, Yansong and Mahon, Kevin and Sun, Yu},
  journal={arXiv preprint arXiv:2411.14256},
  year={2024}
}

@inproceedings{wang2023driveanywheregeneralizableendtoend,
  author={Wang, Tsun-Hsuan and Maalouf, Alaa and Xiao, Wei and Ban, Yutong and Amini, Alexander and Rosman, Guy and Karaman, Sertac and Rus, Daniela},
  booktitle={2024 IEEE International Conference on Robotics and Automation (ICRA)}, 
  title={Drive Anywhere: Generalizable End-to-end Autonomous Driving with Multi-modal Foundation Models}, 
  year={2024},
  pages={6687-6694},
  doi={10.1109/ICRA57147.2024.10611590}
}

@inproceedings{dosovitskiy2017carlaopenurbandriving,
  title={CARLA: An open urban driving simulator},
  author={Dosovitskiy, Alexey and Ros, German and Codevilla, Felipe and Lopez, Antonio and Koltun, Vladlen},
  booktitle={Conference on robot learning},
  pages={1--16},
  year={2017},
  organization={PMLR}
}

@inproceedings{wang2025comdrivecomfortorientedendtoendautonomous,
  title={ComDrive: Comfort-Oriented End-to-End Autonomous Driving},
  author={Wang, Junming and Zhang, Xingyu and Xing, Zebin and Gu, Songen and Guo, Xiaoyang and Hu, Yang and Song, Ziying and Zhang, Qian and Long, Xiaoxiao and Yin, Wei},
  booktitle={2025 IEEE/RSJ International Conference on Intelligent Robots and Systems (IROS)},
  pages={2682--2689},
  year={2025},
  organization={IEEE}
}

@inproceedings{Yao2024ImprovingOG,
  title={Improving out-of-distribution generalization of trajectory prediction for autonomous driving via polynomial representations},
  author={Yao, Yue and Yan, Shengchao and Goehring, Daniel and Burgard, Wolfram and Reichardt, Joerg},
  booktitle={2024 IEEE/RSJ International Conference on Intelligent Robots and Systems (IROS)},
  pages={488--495},
  year={2024},
  organization={IEEE}
}

@inproceedings{guo2025metdrive,
  title={METDrive: Multimodal End-to-End Autonomous Driving with Temporal Guidance},
  author={Guo, Ziang and Lin, Xinhao and Yagudin, Zakhar and Lykov, Artem and Wang, Yong and Li, Yanqiang and Tsetserukou, Dzmitry},
  booktitle={2025 IEEE International Conference on Robotics and Automation (ICRA)},
  pages={6027--6032},
  year={2025},
  organization={IEEE}
}

@inproceedings{chen2020end,
  title={End-to-end autonomous driving perception with sequential latent representation learning},
  author={Chen, Jianyu and Xu, Zhuo and Tomizuka, Masayoshi},
  booktitle={2020 IEEE/RSJ International Conference on Intelligent Robots and Systems (IROS)},
  pages={1999--2006},
  year={2020},
  organization={IEEE}
}

@article{pomerleau1988alvinn,
  title={Alvinn: An autonomous land vehicle in a neural network},
  author={Pomerleau, Dean A},
  journal={Advances in neural information processing systems},
  volume={1},
  year={1988}
}

@InProceedings{Li_2024_CVPR,
    author    = {Li, Zhiqi and Yu, Zhiding and Lan, Shiyi and Li, Jiahan and Kautz, Jan and Lu, Tong and Alvarez, Jose M.},
    title     = {Is Ego Status All You Need for Open-Loop End-to-End Autonomous Driving?},
    booktitle = {Proceedings of the IEEE/CVF Conference on Computer Vision and Pattern Recognition (CVPR)},
    month     = {June},
    year      = {2024},
    pages     = {14864-14873}
}

@ARTICLE{8957584,
  author={Amini, Alexander and Gilitschenski, Igor and Phillips, Jacob and Moseyko, Julia and Banerjee, Rohan and Karaman, Sertac and Rus, Daniela},
  journal={IEEE Robotics and Automation Letters}, 
  title={Learning Robust Control Policies for End-to-End Autonomous Driving From Data-Driven Simulation}, 
  year={2020},
  volume={5},
  number={2},
  pages={1143-1150},
  doi={10.1109/LRA.2020.2966414}}

@ARTICLE{11373828,
  author={Du, Jiayuan and Song, Yuebing and Pan, Xianghui and Su, Shuai and Yang, Jingwei and Wang, Liuyi and Liu, Chengju and Chen, Qijun},
  journal={IEEE Robotics and Automation Letters}, 
  title={BEVDrive-E2E: Imitation With Bird's Eye View Perception for Interpretable End-to-End Autonomous Driving}, 
  year={2026},
  volume={11},
  number={4},
  pages={4353-4360},
  doi={10.1109/LRA.2026.3662561}}

@ARTICLE{11393633,
  author={Syamil, Abi Rahman and Lim, Joonhee and Kum, Dongsuk},
  journal={IEEE Robotics and Automation Letters}, 
  title={NPPC: Neural Parametric Planning Cost for End-to-End Autonomous Driving}, 
  year={2026},
  volume={11},
  number={4},
  pages={4441-4448},
  doi={10.1109/LRA.2026.3663819}}

@ARTICLE{perlad,
  author={Gao, Yinfeng and Zhang, Qichao and Liu, Deqing and Xia, Zhongpu and Li, Guang and Ma, Kun and Chen, Guang and Ye, Hangjun and Chen, Long and Ding, Da-Wei and Zhao, Dongbin},
  journal={IEEE Robotics and Automation Letters}, 
  title={PerlAD: Towards Enhanced Closed-Loop End-to-End Autonomous Driving With Pseudo-Simulation-Based Reinforcement Learning}, 
  year={2026},
  volume={11},
  number={5},
  pages={5821-5828},
  doi={10.1109/LRA.2026.3675928}}

@ARTICLE{MetaTraj,
  author={Shi, Xiaodan and Zhang, Haoran and Yuan, Wei and Shibasaki, Ryosuke},
  journal={IEEE Transactions on Intelligent Transportation Systems}, 
  title={MetaTraj: Meta-Learning for Cross-Scene Cross-Object Trajectory Prediction}, 
  year={2023},
  volume={24},
  number={12},
  pages={14000--14009},
  doi={10.1109/TITS.2023.3299112}}

@ARTICLE{9700770,
  author={Le Mero, Luc and Yi, Dewei and Dianati, Mehrdad and Mouzakitis, Alexandros},
  journal={IEEE Transactions on Intelligent Transportation Systems}, 
  title={A Survey on Imitation Learning Techniques for End-to-End Autonomous Vehicles}, 
  year={2022},
  volume={23},
  number={9},
  pages={14128--14147},
  doi={10.1109/TITS.2022.3144867}}

@ARTICLE{Multimodal,
  author={Xiao, Yi and Codevilla, Felipe and Gurram, Akhil and Urfalioglu, Onay and López, Antonio M.},
  journal={IEEE Transactions on Intelligent Transportation Systems}, 
  title={Multimodal End-to-End Autonomous Driving}, 
  year={2022},
  volume={23},
  number={1},
  pages={537--547},
  doi={10.1109/TITS.2020.3013234}}

\begin{IEEEbiography}[{\includegraphics[width=1in,height=1.25in,clip,keepaspectratio]{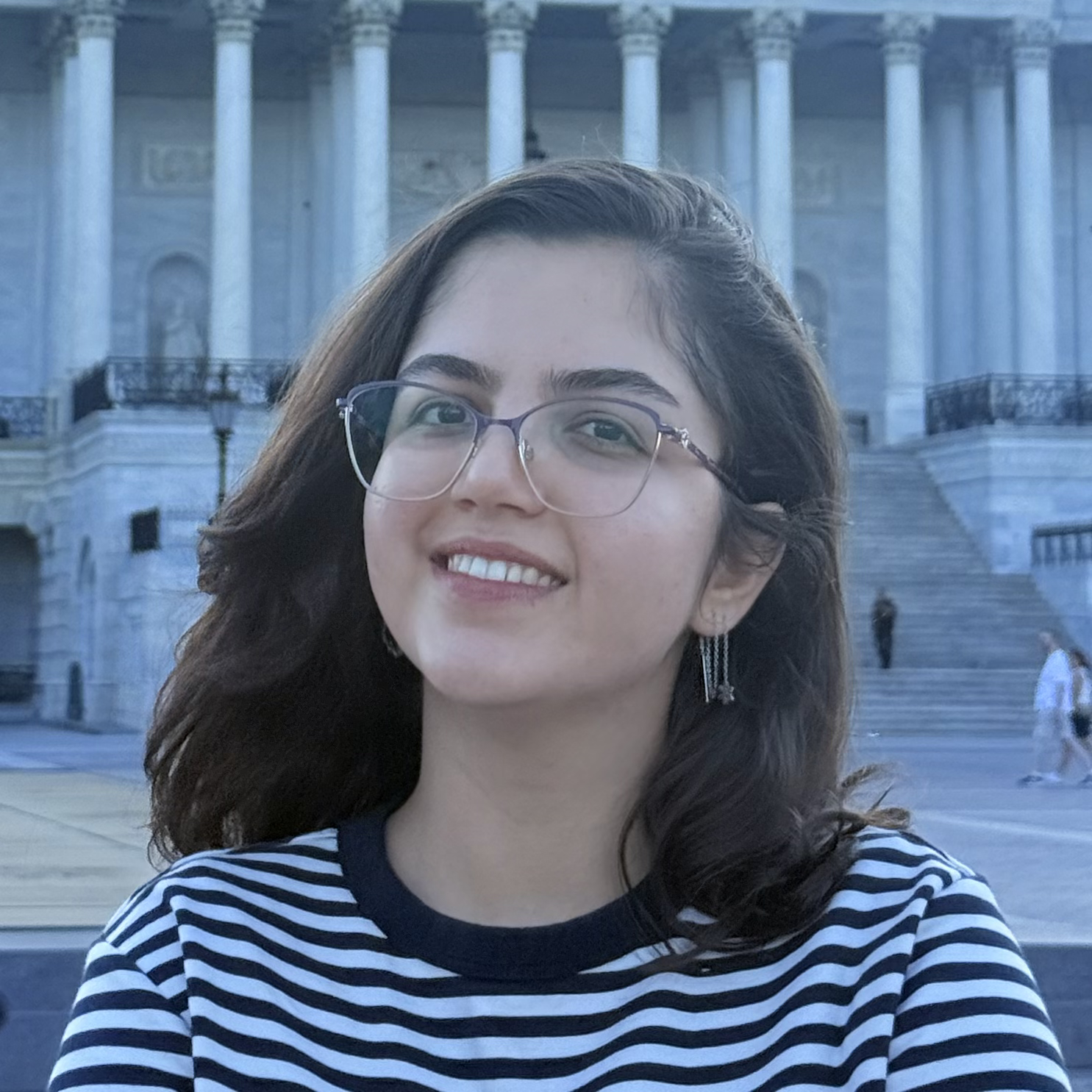}}]{Fatemeh Naeinian}
received the B.Sc. degree in electrical engineering, with a minor in computer
engineering, from the University of Tehran, Tehran, Iran. She is currently
pursuing the Ph.D. degree in electrical engineering at New York University,
New York, NY, USA, under the supervision of Prof. Anna Choromanska. Her
research interests include self-supervised learning, representation learning,
world models, and robust generalization in end-to-end autonomous driving.
\end{IEEEbiography}

\begin{IEEEbiography}[{\includegraphics[width=1in,height=1.25in,clip,keepaspectratio]{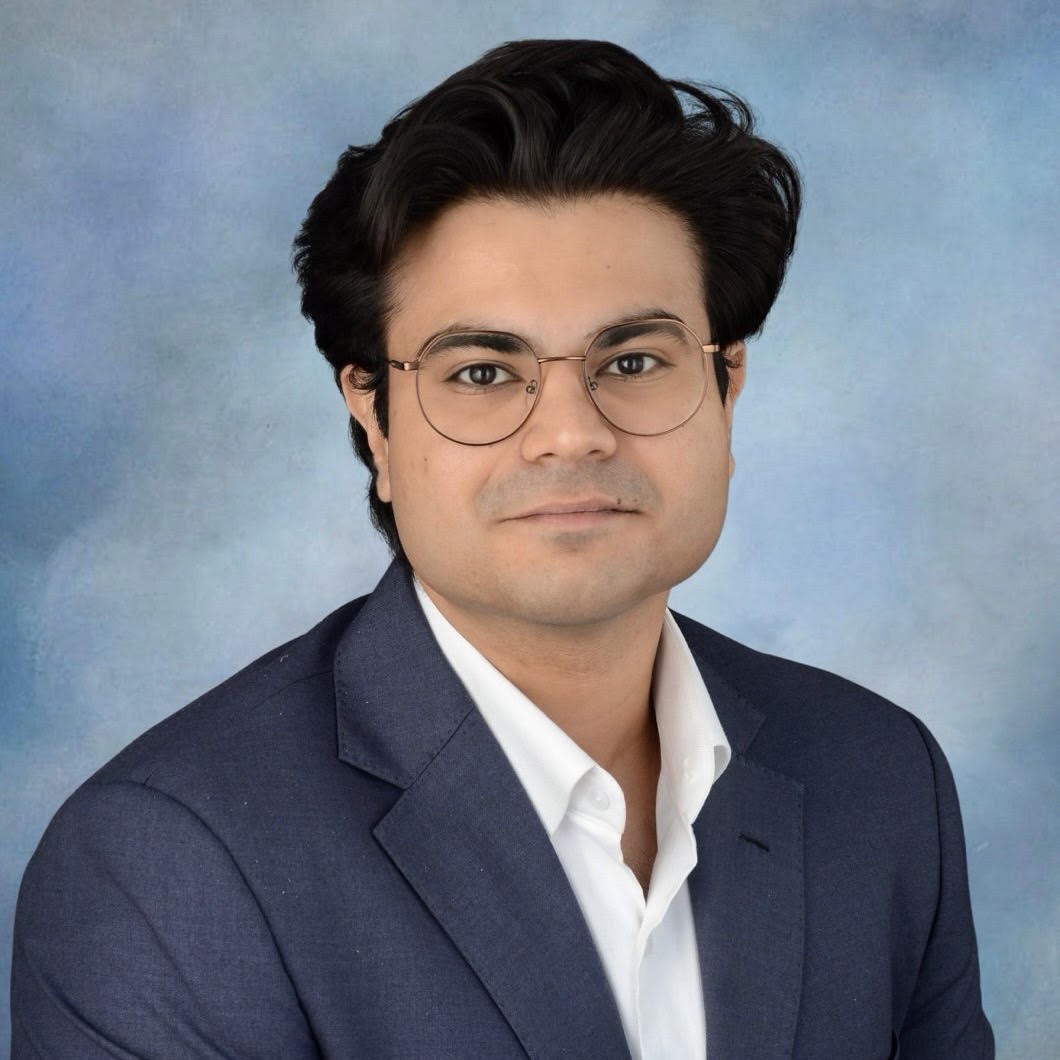}}]{Ali Hamza}
received the B.S. degree in computer science from Habib University, Karachi,
Pakistan, in 2022, and the M.S. degree in computer engineering from the NYU
Tandon School of Engineering, New York University, New York, NY, USA, in
2026. His research interests include machine learning, self-supervised
learning, computer vision, and autonomous driving.
\end{IEEEbiography}

\begin{IEEEbiography}[{\includegraphics[width=1in,height=1.25in,clip,keepaspectratio]{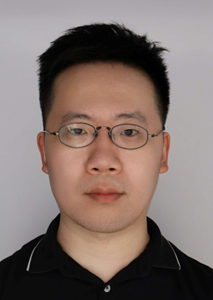}}]{Haoran Zhu}
received the Ph.D. degree from New York University, New York, NY, USA, where
he worked on deep learning for autonomous driving under the supervision of
Prof. Anna Choromanska. He is currently a Postdoctoral Researcher at New York
University. His research
interests include world models, self-supervised learning, end-to-end learning,
and intelligent systems for the physical world.
\end{IEEEbiography}

\begin{IEEEbiography}[{\includegraphics[width=1in,height=1.25in,clip,keepaspectratio]{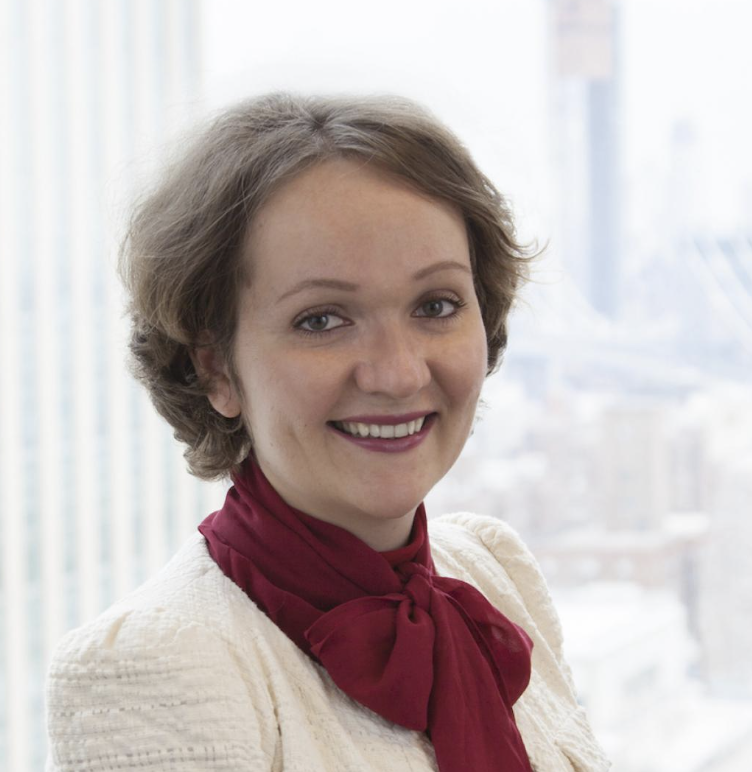}}]{Anna Choromanska}
received the M.Sc. degree from Warsaw University of Technology, Warsaw,
Poland, in 2009, and the M.Phil. and Ph.D. degrees in electrical engineering
from Columbia University, New York, NY, USA, in 2014. She is currently an
Associate Professor with the Department of Electrical and Computer Engineering,
NYU Tandon School of Engineering, New York University, and Director of the
Learning Systems Laboratory. Her research interests include deep learning,
optimization, self-supervised learning, Joint Embedding Predictive Architectures,
world models, and autonomous driving.
\end{IEEEbiography}

\end{document}